\pdfoutput=1
\documentclass{article} %
\usepackage{iclr2027_conference,times}

\newif\ifarxiv
\arxivtrue
\ifarxiv\iclrfinalcopy\fi

\usepackage{amsmath,amsfonts,bm}

\def\eqref#1{equation~\ref{#1}}

\def\1{\bm{1}}

\DeclareMathAlphabet{\mathsfit}{\encodingdefault}{\sfdefault}{m}{sl}
\SetMathAlphabet{\mathsfit}{bold}{\encodingdefault}{\sfdefault}{bx}{n}

\usepackage{hyperref}
\usepackage{url}
\ifarxiv
\hypersetup{pdftitle={What Enables In-Context Behavior Prompting for Manipulation?},
            pdfauthor={Austin Patel, Ben Pekarek, Joel Enrique Castro Hernandez, Shuran Song}}
\fi

\usepackage{xcolor}
\ifarxiv\hypersetup{colorlinks=true, linkcolor={red!65!black}, citecolor={blue!70!black}, urlcolor={blue!70!black}}\fi
\usepackage{soul}
\usepackage{graphicx}
\usepackage{xspace}
\usepackage{wrapfig}
\usepackage{siunitx}
\usepackage{float}
\usepackage{multirow}
\usepackage{enumitem}
\usepackage{pifont}
\usepackage{fancyvrb}
\usepackage{fvextra}

\newcommand{\cmark}{\ding{51}}  %
\newcommand{\xmark}{\ding{55}}  %

\definecolor{alg_color}{HTML}{228B22} %
\definecolor{data_color}{HTML}{0047AB} %
\definecolor{eval_color}{HTML}{CC5500} %
\newcommand{\ours}{BPP\xspace}
\newcommand{\goalimage}{\texttt{Goal-Image}\xspace}
\newcommand{\lang}{\texttt{Language}\xspace}
\newcommand{\icrt}{ICRT\xspace}
\newcommand{\liberogen}{LIBERO-Gen\xspace}
\newcommand{\liberogencombination}{\liberogen Combination\xspace}
\newcommand{\liberogenchain}{\liberogen Chain\xspace}
\newcommand{\pizerofive}{$\pi_{0.5}$\xspace}

\usepackage{etoolbox}
\makeatletter
\patchcmd{\@outputpage}{\vfil\color@hbox}{\vfil\pdfrunninglinkoff\color@hbox}{}{\PackageWarning{main}{head patch failed}}
\patchcmd{\@outputpage}{\color@endbox}{\color@endbox\pdfrunninglinkon}{}{\PackageWarning{main}{head-end patch failed}}
\patchcmd{\@outputpage}{\baselineskip\footskip\color@hbox}{\baselineskip\footskip\pdfrunninglinkoff\color@hbox}{}{\PackageWarning{main}{foot patch failed}}
\makeatother

\title{What Enables In-Context Behavior Prompting for Manipulation?}

\ifarxiv  %
\author{Austin Patel$^{1}$ \quad Ben Pekarek$^{1}$ \quad Joel Enrique Castro Hernandez$^{2}$ \quad Shuran Song$^{1}$ \\[2pt]
$^{1}$Stanford University \quad $^{2}$University of California, Berkeley \\[2pt]
{\hypersetup{pdfborder={0 0 0}}\href{https://behavior-prompting.github.io}{\textcolor{blue}{behavior-prompting.github.io}}}
}
\else  %
\author{Austin Patel$^{1}$, Ben Pekarek$^{1}$, Joel Enrique Castro Hernandez$^{2}$ \& Shuran Song$^{1}$ \\
$^{1}$Stanford University \quad $^{2}$University of California, Berkeley \\
\href{https://behavior-prompting.github.io}{behavior-prompting.github.io}
}
\fi

\begin{document}

\maketitle
\ifarxiv\lhead{}\renewcommand{\headrulewidth}{0pt}\fi
\ifarxiv\vspace{-\baselineskip}\vspace{-4pt}\fi  %

\begin{abstract}

Behavior prompting is a paradigm in which a sensorimotor robot demonstration, called a behavior prompt, serves as an in-context prompt for performing new tasks at test time. While prior work has shown that this capability is possible, the conditions that enable it remain poorly understood. We present an empirical study of when, how, and why behavior prompting works. To support this study, we introduce DrawAnything and \liberogen, benchmarks with up to 2000 procedurally generated tasks that evaluate test-time adaptation to unseen drawing and tabletop manipulation tasks. We also present Behavior Prompting Policy (BPP), an in-context visuomotor architecture, and iPhUMI, a handheld interface to demonstrate behavior prompts at test time. Our main finding is that task diversity, rather than demonstrations per task, is a key driver of prompting capability. Given sufficient diversity, a behavior prompt improves adaptation to unseen tasks, reducing drawing error by 80.7\% over goal-image conditioning and improving success on chained manipulation tasks by up to 20.8\% over language conditioning. Given insufficient diversity in a real-world laundry experiment, behavior prompting has weaker task conditioning than a language baseline. An attention analysis shows how the prompt is used: the policy follows it step by step as a source of dense sub-goals. Prompt ablations show that dense sensorimotor detail matters: removing actions or downsampling the prompt hurts fine-grained action adaptation. We have open-sourced all components to enable reproducible research on behavior prompting without needing industrial-scale data collection or compute.%

\end{abstract}

\begin{figure}[h]
    \centering
    \vspace{-2mm}
    \includegraphics[width=\linewidth]{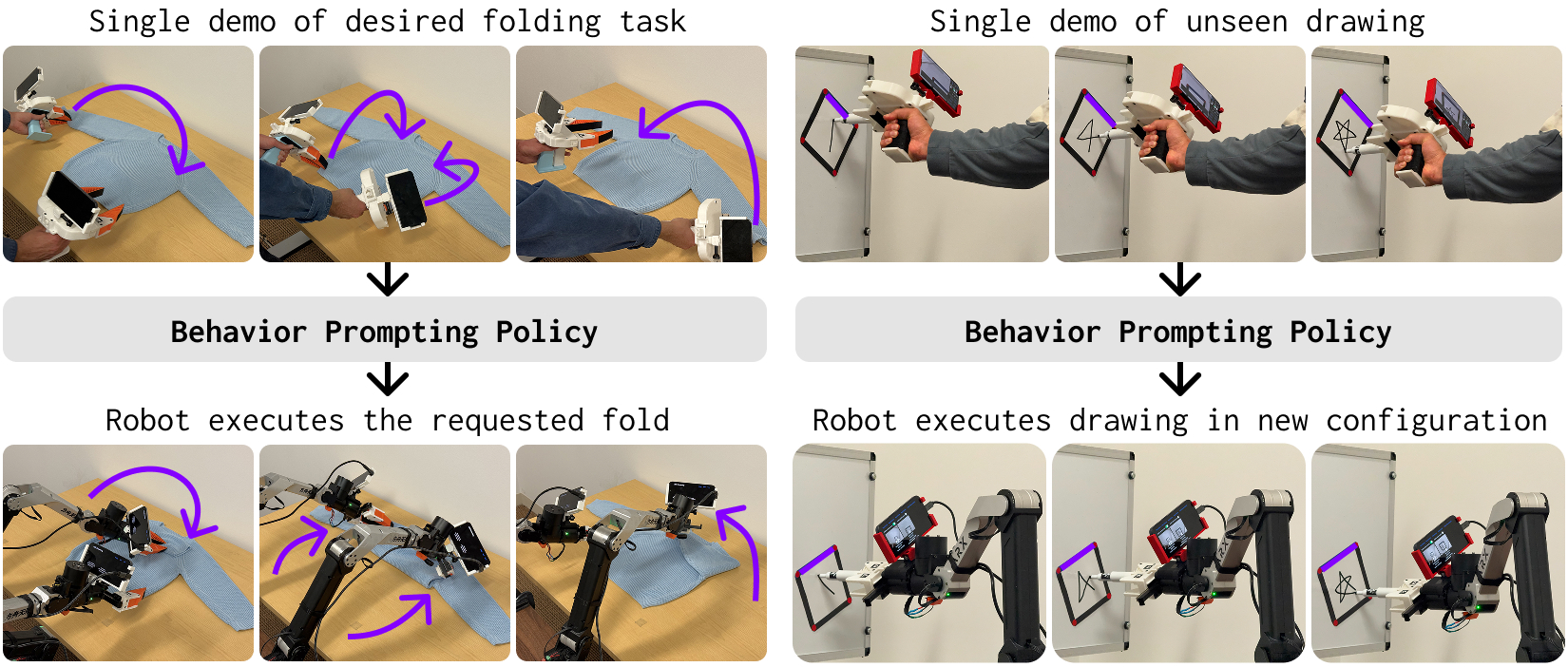}
    \vspace{-6.5mm}
    \caption{\textbf{Behavior prompting} conditions test-time execution on a single robot demonstration. This enables a user to specify a task via demonstration (left) or define new robot capabilities (right).}
    \vspace{-4mm}
    \label{fig:teaser}
\end{figure}

\section{Introduction}

Teaching a robot a new skill typically requires fine-tuning a policy~\citep{black2026pi0visionlanguageactionflowmodel,kim2025fine}. Large language models instead adapt to new tasks from a few examples placed in context, with no parameter updates~\citep{NEURIPS2020_1457c0d6}. \textit{Behavior prompting} brings this paradigm to manipulation (Fig.~\ref{fig:teaser}): at test time, the policy is conditioned in-context on a single demonstration of the desired task, consisting of observations and actions in the robot's sensorimotor space (i.e., a \textit{behavior prompt}). This prompt specifies \textit{what} the task is and an example of \textit{how} to complete it, whereas language instructions often leave the \textit{how} underspecified~\citep{gu2024rttrajectory,jiang2023vima}.

Adapting a policy from a demonstration has a long history in few-shot imitation learning~\citep{duan2017one,finn2017one,james2018task,mandi2022}. More recently, ICRT~\citep{fu2024icrt} showed that an in-context policy can condition directly on raw sensorimotor demonstrations (i.e., behavior prompts) to execute new tasks at test time. ICRT establishes that behavior prompting is possible, but what remains missing is an understanding of the conditions that enable it.

\textbf{Our goal is to understand what enables behavior prompting: \textit{when} it works, \textit{how} the policy uses the prompt, and \textit{why} the prompt helps}. Such a study needs a setting where training task diversity can be varied and where new \textit{unseen} tasks, specified at test time, require the policy to rely on the prompt. It also needs ablations cheap enough to run systematically. ICRT's evaluation, a single real-world tabletop domain with six motion primitives, was designed to validate the approach rather than to support such a study, since both its task diversity and its ablations are limited by the cost of real-world data collection and evaluation. Existing simulation benchmarks remove this cost and enable reproducibility, but their scenes often reveal the task~\citep{yu2020meta,james2019rlbench}, making the prompt unnecessary, or, like LIBERO~\citep{liu2023libero}, only evaluate on training tasks.

We therefore study three manipulation domains across simulation and the real world. \textbf{DrawAnything} (simulation and real) trains on up to 2000 procedurally generated drawings and tests whether a policy can reproduce new human-drawn figures at test time, which requires fine-grained action adaptation. \textbf{\liberogen} (simulation) procedurally extends LIBERO to hundreds of tabletop manipulation tasks within a shared scene to test adaptation to new combinations and sequences of skills. \textbf{Laundry folding} (real) probes the low-diversity regime with only three tasks. To study them, we introduce Behavior Prompting Policy (BPP), a visuomotor architecture for behavior prompting that cross-attends to an encoded prompt and generates actions with diffusion~\citep{chi2023diffusionpolicy}. For the real-world experiments, we design iPhUMI, an iPhone-based handheld gripper building on UMI~\citep{chi2024universal} that lets a human specify behavior prompts at test time without a robot.

Together, these let us answer three questions that were previously out of reach. \textbf{When does behavior prompting work?} Ablating the training data, we find that task diversity, rather than demonstrations per task, is a key driver of prompting capability. With sufficient diversity, a behavior prompt improves adaptation to unseen tasks, reducing drawing error by 80.7\% over goal-image conditioning and improving success on chained tabletop manipulation by up to 20.8\% over language conditioning. In our laundry folding experiment where task diversity is low, task conditioning weakens: the policy identifies the intended fold less reliably through demonstration than with language. \textbf{How does the policy use the prompt?} Analyzing the policy's attention, we find that it tracks the part of the prompt matching its current progress, using it as a source of dense sub-goals. \textbf{Why does the prompt help?} Ablating the prompt contents on DrawAnything, we find that the advantage comes from dense sensorimotor detail: observations anchor the lookup into the prompt, and removing actions or downsampling the prompt below 1\,Hz degrades fine-grained action adaptation.

In summary, our contributions are:
\textbf{1)}~DrawAnything and \liberogen, benchmarks with up to 2000 procedurally generated tasks that evaluate behavior prompting on unseen tasks.
\textbf{2)}~An empirical study of when, how, and why behavior prompting works, with findings on task diversity, how the policy uses the prompt, and which prompt contents matter.
\textbf{3)}~Behavior Prompting Policy, a strong reference architecture for prompting.
We have open-sourced all components to enable reproducible research on behavior prompting without needing industrial-scale data collection or compute.

\section{Related Work}
\label{sec:related_work}

\noindent\textbf{Test-time Adaptation to New Tasks.} Language-conditioned multi-task policies~\citep{jang2021bc,octo_2023} and VLA models~\citep{rt22023arxiv,kim24openvla,black2026pi0visionlanguageactionflowmodel,intelligence2025pi05visionlanguageactionmodelopenworld} achieve zero-shot adaptation to some unseen environments and objects, but rarely to new low-level actions or skills. Few-shot methods instead adapt to a new task from a small set of demonstrations, and differ in how they use the demonstration. \textit{Weight-update} methods adapt parameters with a few gradient steps via meta-learning~\citep{finn2017one,Yu-RSS-18} or online via test-time training~\citep{jiang2026robottt0}. \textit{Structured} methods transfer a demonstration to a new scene through explicit reference frames (e.g., task-parameterized models~\citep{Calinon16JIST} and align-then-replay~\citep{valassakis2022dome}), learned correspondences~\citep{zhang2024one}, or task priors~\citep{doi:10.1126/scirobotics.adv7594}, trading generality for sample efficiency. \textit{Conditioning} methods keep the weights fixed and feed the demonstration to the policy, either compressed into a task embedding~\citep{james2018task} or attended to in context~\citep{duan2017one,mandi2022,Jain-2024-144319,xu2022prompt,shah2025mimicdroid}, in some cases retrieved at test time from a larger set~\citep{sridhar2025ricl,papagiannis2025rplusx}.

In-context few-shot methods further differ in the prompt contents. Rather than raw observations and actions, many represent the demonstration as keypoints~\citep{dipalo2024kat}, graphs~\citep{vosylius2024instant}, multi-modal prompts~\citep{jiang2023vima}, reasoning traces~\citep{nguyen2026iclr}, or structured rationales~\citep{xu2026stellavlaincontextstructureddemonstration}. Others prompt with human video~\citep{Yu-RSS-18,Jain-2024-144319,wang2023mimicplay,xu2023xskill,kim2025uniskill,zhu2026learninggeneralizablerobotpolicy,zhou2026zerowam,jiang2026robottt0,rhoda2026dva}, which is easier to collect but crosses an embodiment gap.

\noindent\textbf{Behavior Prompting.} 
Behavior prompting is an in-context few-shot conditioning method in which the prompt is a raw demonstration in the robot's sensorimotor space. Compared to human video, \textit{a behavior prompt has no embodiment gap and is already accessible from existing demonstration data during training}. However, performing a new task requires one test-time demo in the robot's space, e.g., with a handheld gripper. ICRT~\citep{fu2024icrt} introduces a behavior prompting architecture and validates the paradigm on a real robot in a single tabletop domain, training on 1098 trajectories spanning 29 tasks and 6 motion primitives. ICRT ablates model initialization, DROID pretraining, and the training loss, but does not study training task diversity, prompt modalities, or how the policy attends to the prompt, all of which we study. Concurrent industrial systems also report in-context demonstration prompting, where the capability either emerges from long-context training~\citep{generalist2026gen15} or is trained for explicitly~\citep{skild2026s1}, attributing it to pretraining scale and to task diversity, respectively. \textbf{\textit{These works show that such prompting is possible but none isolates the enabling factors in a controlled, reproducible study, which is the focus of this work.}}

Existing benchmarks do not support such a study out of the box. They evaluate only on training tasks~\citep{liu2023libero,zhou2025liberopro,fei25libero-plus}, in LIBERO's case language-sufficient and performance-saturated (Tab.~\ref{tab:libero_results_std}); offer tasks identifiable from the scene alone, so the prompt is unnecessary~\citep{yu2020meta,james2019rlbench}; or provide no per-task demonstrations~\citep{yang2026robolab}. Closest to our setting, VIMA-Bench~\citep{jiang2023vima} holds out novel prompt-specified tasks, but over object rearrangements with fixed action primitives, and its prompts are object-centric rather than sensorimotor. Data generators such as RoboTwin 2.0~\citep{chen2025robotwin} could be extended to host our protocol, just as we extended LIBERO into \liberogen. Our benchmarks combine many demonstrable tasks per scene and support specifying new tasks at test time via a sensorimotor demonstration; they allow researchers to meaningfully study algorithm design of behavior prompting without the requirements of industrial-scale data and compute.

\section{Method}
\label{sec:method}

\subsection{\textbf{Behavior Prompt Definition}}
\label{sec:what_is_behavior_prompt}

A behavior prompt is a single demonstration of a desired task (at a different environment configuration) consisting of a sequence of observations, proprioception, and actions in the robot's sensorimotor execution space.
At test time, the prompt is a training demonstration or, for new tasks, a human demonstration (e.g., with a handheld gripper). A demonstration is most useful when it clarifies what language leaves underspecified: spatial-temporal structure (e.g., a drawing), preferences easier to show than describe (e.g., a preferred fold), or a manipulation strategy (e.g., a grasp type). \textit{Behavior prompting thus inherits the benefits of video conditioning (i.e., visual sub-goals for the task) while adding actions}, which we find improve fine-grained action adaptation (e.g., following drawings).

\subsection{\textbf{Behavior Prompting Policy (BPP)}}
\label{sec:prompting_policy}

\begin{figure*}[t]
    \centering
    \includegraphics[width=1\linewidth]{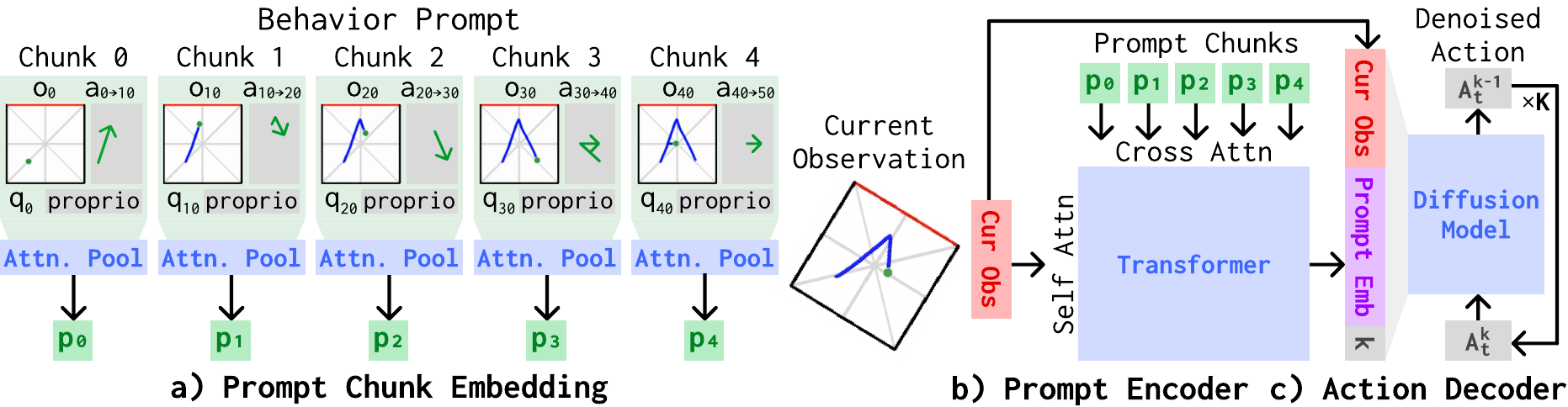} %
    \vspace{-5mm}
    \caption{\textbf{Behavior Prompting Policy architecture.} a) Every $\Delta t$ steps of the behavior prompt form a chunk that contains one step of observation and proprio along with $\Delta t$ actions. Attention pooling merges $\{o,q,a\}$ into a chunk embedding $p_i$. The policy consists of: b) a prompt encoder, which extracts relevant prompt information given the current obs, and c) an action decoder, which reasons over the current obs and relevant prompt information to generate actions over $K$ diffusion steps.}

    \label{fig:prompting_architecture}
    \vspace{-3mm}
\end{figure*}

Behavior Prompting Policy is an in-context visuomotor architecture that takes a behavior prompt and current observation as input and outputs closed-loop actions (Fig.~\ref{fig:prompting_architecture}). BPP consists of a prompt encoder and an action decoder, like in Vid2Robot~\citep{Jain-2024-144319}.   We position BPP as a strong reference architecture for behavior prompting, with practical benefits, discussed below.

\clearpage

\textbf{Prompt Encoder~(Fig. \ref{fig:prompting_architecture}b):}
To encode the prompt, we temporally downsample the observations and proprioception for computational efficiency (typically to 1Hz). Actions are not downsampled to retain the full behavior sequence. The prompt is a sequence of chunks each consisting of a single timestep of observation $o$, proprioception $q$, and the sequence of actions $a$ leading up to the next chunk. We apply attention pooling per chunk to merge $\{o,q,a\}$ into a single chunk embedding $p_i$ forming the sequence $P = [p_0, p_1, \ldots, p_n]$ where $n$ varies with prompt length. The pooling temporally associates information from the same timestep and reduces the prompt sequence length.

The prompt encoder is a transformer decoder that extracts relevant prompt information given the current observation. We tokenize the current observation by representing each observation entry with a single token per timestep of history and then perform cross-attention with the prompt chunk embeddings $P$. We use learned positional embeddings for the prompt and the current observation.

\textbf{Action Decoder~(Fig. \ref{fig:prompting_architecture}c):}
Given the relevant prompt information extracted by the prompt encoder, the policy now needs to generate actions. We concatenate the current observation, the extracted prompt information, and diffusion timestep $k$ and pass it to a diffusion model to denoise actions. We use the CNN action diffusion architecture with FiLM~\citep{perez2018film} from~\citet{chi2023diffusionpolicy}.

\textbf{Training:}
We follow standard behavior cloning training practices for action diffusion policies with additional care taken to handle the behavior prompt. For each training step, we sample a single demonstration from the training data as the prompt. We then load a batch of receding horizon observations and future action chunks from other demonstrations of the \textit{same task}. Each demonstration will have varying environment configurations, so the policy must reason about the \textit{temporal similarities} and \textit{spatial differences} between the prompt and the current observation to generate actions.

\textit{\textbf{During training we require no explicit correspondence, spatially or temporally, between two demonstrations of the same task}}. The prompt is simply provided as an additional model input and the model learns end-to-end how to leverage it. Since prompts are demonstrations, \textit{\textbf{BPP can be directly trained on existing multi-task imitation datasets with no additional data collection}}.

\textbf{Inference:} We select a single prompt per rollout, encode it once, and cache the chunk embeddings.

\textbf{Comparison to ICRT:} BPP differs from ICRT~\citep{fu2024icrt} in three ways: (i) prompt handling, cross-attention from a separate encoder versus causal self-attention; (ii) the action decoder, diffusion versus L1; and (iii) history, a short window versus the full rollout retained in context. \S\ref{sec:when} isolates the benefit of (ii) and the effect of (iii) under disturbance. Two practical benefits follow: action diffusion accounts for most of BPP's improvement over ICRT, and BPP's short window never reaches a context limit during execution, unlike ICRT's retained history. Further comparisons are in \S\ref{sec:bpp_comparison_to_prior_methods}.

\subsection{\textbf{iPhUMI}}
\label{sec:iphumi}

iPhUMI is a handheld gripper for collecting training data and specifying behavior prompts. It adapts the UMI gripper~\citep{chi2024universal} by replacing the GoPro with an iPhone, following prior iPhone-based grippers~\citep{shafiullah2023bringing}. ARKit provides real-time localization, removing the mapping stage, and our app wirelessly streams a demo to prompt BPP at test time. See \S\ref{sec:appendix_iphumi} for details.

\clearpage

\section{Evaluation}
\label{sec:evaluation}

\begin{figure*}[t]
    \centering
    \includegraphics[width=\linewidth]{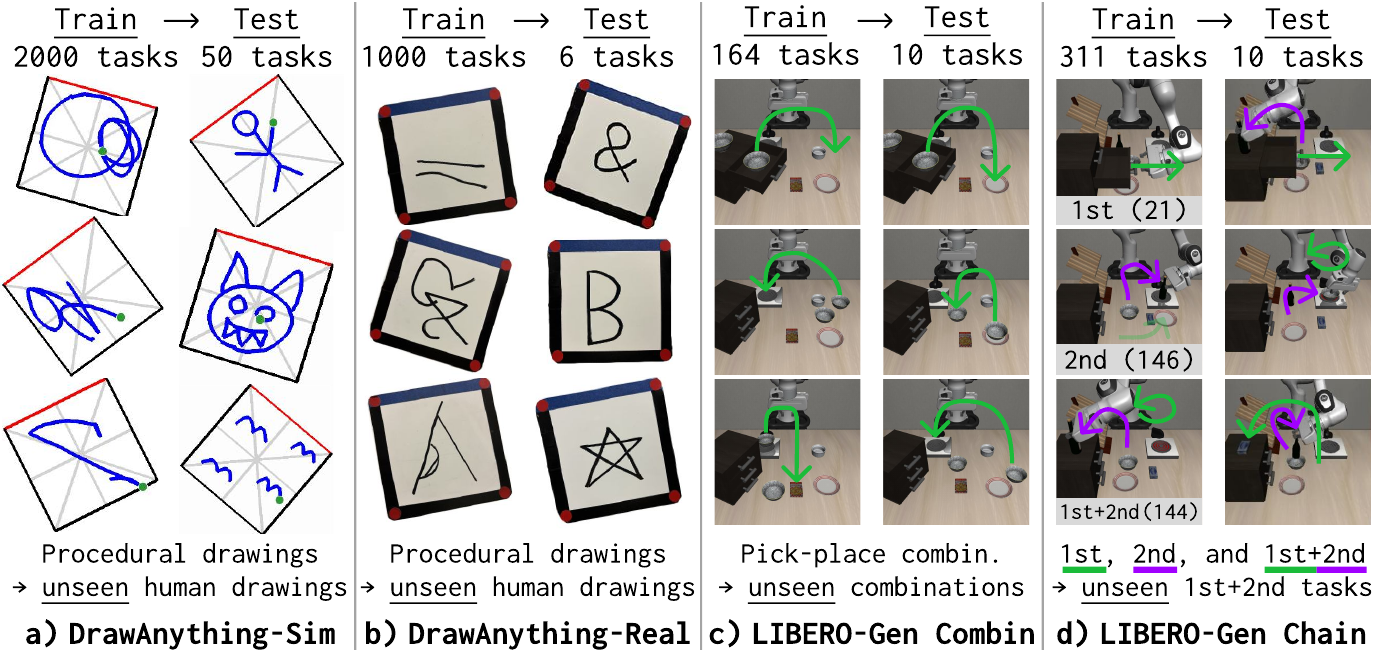}
    \vspace{-4mm}
    \caption{\textbf{Benchmarking Suite.} In DrawAnything (a, b), we evaluate whether a policy can recreate a previously unseen drawing at varying board poses given a single human demo. In \liberogencombination (c) two identical bowls are randomly positioned, and the robot is given instructions for which one to grasp and where to place it. In \liberogenchain (d) we explore the set of two step interactions. We have: 1) \textit{first step tasks} include open middle/top drawer, push plate, turn on stove, and pick-place, 2) \textit{second step tasks} just do second action (pick-place only) with first action already done, and 3) \textit{chained tasks} where we do a first step task then a second step task in succession.}
    \label{fig:tasks}
    \vspace{-3mm}
\end{figure*}

We introduce DrawAnything and \liberogen (Fig.~\ref{fig:tasks}), benchmarks with high task diversity, procedurally generated demonstrations, and test-time task specification of new tasks.

\begin{itemize}[leftmargin=*]
    \item \textbf{DrawAnything-Sim~(Fig.~\ref{fig:tasks}a):} Drawing is well suited for behavior prompting as it requires the policy to continuously reference low-level drawing instructions in the prompt.
    We train on 2000 procedurally generated drawing tasks with 5 demos per task at randomized board orientations. We evaluate on 50 unseen drawing tasks collected by a human to ensure significant variation from the training tasks. During evaluation, the policy must reconstruct a previously unseen drawing given a single human demo. The rollout occurs at a different board orientation than the prompt.

    \vspace{1mm}

    \item \textbf{DrawAnything-Real~(Fig.~\ref{fig:tasks}b):} 
    We create an analogous real-world drawing setup using an ARX robot arm with an iPhone wrist camera and marker attached to the end effector (Fig.~\ref{fig:teaser}). Within a large whiteboard we have a square drawing region that can be placed at any position and in a \ang{90} rotation range. We collect 1000 training tasks (200 tasks at 5 demos/task from a human using iPhUMI and 800 tasks at 6 demos/task with a scripted policy). Evaluation is done on 10 tasks (4 training, 6 unseen) collected by a human using iPhUMI. This benchmark requires full 6DoF action (compared to 2D in sim) and robustness to visual occlusions caused by the marker. 

    \vspace{1mm}

    \item \textbf{\liberogencombination~(Fig.~\ref{fig:tasks}c):} We introduce \liberogen, a procedural generation framework for LIBERO~\citep{liu2023libero}. Compared to other LIBERO extensions that focus on visual or semantic robustness~\citep{fei25libero-plus,zhou2025liberopro}, \liberogen evaluates \textit{test-time instruction-following capability on unseen tasks} by generating new environments, tasks, and demonstrations. Using this tool, we create \liberogencombination which extends the 10 tasks in LIBERO Spatial with 164 more tasks in the same environment. All tasks involve picking one of the two identical bowls and placing it on one of nine locations. For evaluation, we hold out 10 ``combinations'' of pick-place locations seen individually during training, but never jointly.
    
    \vspace{1mm}

    \item \textbf{\liberogenchain~(Fig.~\ref{fig:tasks}d):} Using \liberogen, we create \liberogenchain, which extends the 10 original tasks in LIBERO Goal with 311 additional tasks in the same environment. The tasks involve sequentially executing two single-step tasks in a ``chain.'' To understand whether behavior prompting can perform unseen chains of manipulation skills at test time, we hold out 10 two-step tasks consisting of individually seen manipulation skills.

\end{itemize}

\subsection{When does behavior prompting work?}
\label{sec:when}

\begin{figure*}[t]
    \centering
    \includegraphics[width=\linewidth]{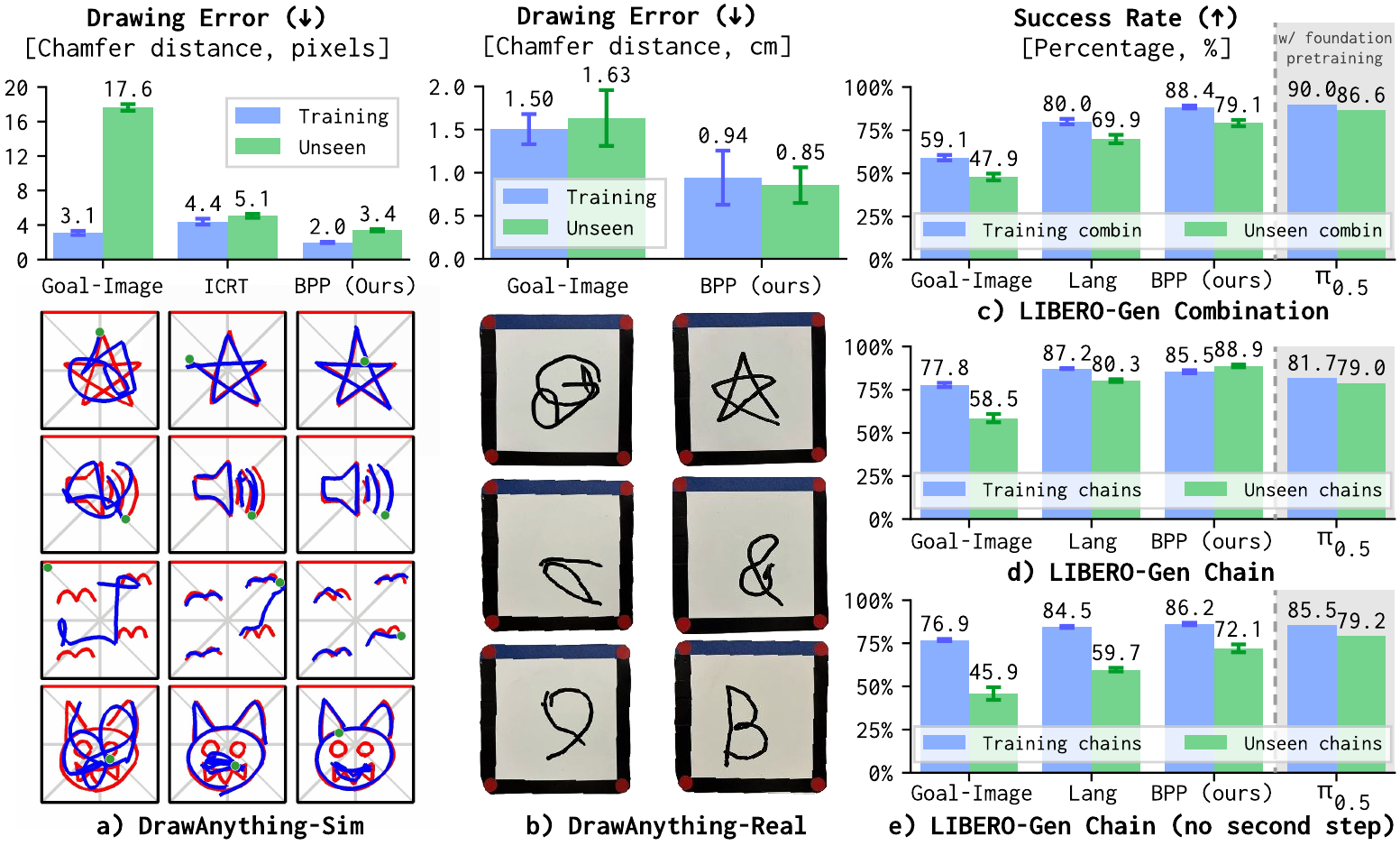}
    \vspace{-6mm}
    \caption{\textbf{Benchmark Results.} For DrawAnything (a,b) we find \goalimage performs well only on training drawings, while BPP and \icrt perform well on unseen drawings. Side-by-side qualitative results in (a,b) are for \textit{unseen} drawings. For unseen manipulation tasks in \liberogen (c,d,e), \ours outperforms baselines and rivals $\pi_{0.5}$ despite not having foundation pretraining. We report the $\pm1$ stdev error bar across three seeds for sim ($\pi_{0.5}$ is single seed) and across tasks for a single seed for the real-world drawing.}
    \label{fig:main_result}
    \vspace{-3mm}
\end{figure*}

We compare to \goalimage and \lang that match the \ours action decoder but replace the prompt encoder with ViT goal image encoding~\citep{dosovitskiy2020image} and finetuned CLIP language encoding~\citep{radford2021learning}, respectively. We also compare to ICRT~\citep{fu2024icrt}, a prior behavior prompt model, and \textit{finetuned} $\pi_{0.5}$~\citep{intelligence2025pi05visionlanguageactionmodelopenworld}, a VLA with foundation pretraining. $\pi_{0.5}$ has foundation pretraining so it is not directly comparable.

\noindent\textbf{\textit{Behavior prompting improves test-time adaptation to unseen tasks.}}
\textsc{DrawAnything:} We observe \ours can reconstruct new drawings from a single human demonstration via mouse in sim or iPhUMI in real (Fig.~\ref{fig:main_result}a,b). For unseen drawings, BPP reduces error compared to \goalimage by 80.7\% in sim and 47.9\% in real. 
A single goal image only indicates \textit{what} drawing to do, while a behavior prompt describes a step-by-step example for \textit{how} to complete a drawing. Against \icrt~\citep{fu2024icrt}, \ours reaches an error of $3.43 \pm 0.10$ versus $5.07 \pm 0.22$, a 32.3\% reduction. To dissect this gain, replacing \ours's diffusion head with L1 regression to match \icrt raises its error to $4.84 \pm 0.22$, so the action decoder accounts for most of the gap. Because \icrt retains the full rollout history, a mid-rollout disturbance that reverts the canvas to an earlier state leaves its context out of distribution: its error increases by $10.47 \pm 0.98$, versus $5.48 \pm 0.15$ for \ours, whose fixed-length history is less affected (disturbance protocol in \S\ref{sec:bpp_comparison_to_prior_methods}; recovery mechanism discussed in \S\ref{sec:how}).

\textsc{\liberogen:} \ours improves test-time adaptation to \textit{unseen} manipulation tasks for both \liberogen benchmarks. For \liberogencombination there are two decisions (which bowl to pick and where to place it) and for \liberogenchain there are up to four (ex1: \textit{pick, place, pick, place}, ex2: \textit{turn stove knob on, pick, place}). Language explicitly indicates these steps, while the goal image only provides the final state. Although language names each step, a behavior prompt additionally grounds each step in the robot's sensorimotor space, showing where to grasp and place and how the steps connect. In \liberogencombination (Fig.~\ref{fig:main_result}c), \ours improves on unseen tasks by 13.2\% over \lang. In \liberogenchain (Fig.~\ref{fig:main_result}d,e) we ablate including 2nd step tasks in the training data (1st,2nd,1st+2nd $\rightarrow$ 1st,1st+2nd). Removing 2nd step tasks makes adaptation to unseen chained tasks (1st+2nd) more challenging as the policy will have never seen the 2nd step after the 1st step has completed (ex: \textit{never seen how to put the wine on the cabinet after a cabinet drawer is opened}). In this ablation, \ours achieves a 20.8\% gain over \lang in Fig.~\ref{fig:main_result}e (without 2nd step) compared to a 10.7\% gain over \lang in Fig.~\ref{fig:main_result}d (with 2nd step); in other words, a larger unseen adaptation gap yields greater \ours gains. Additionally, despite having no pretraining, \ours rivals finetuned $\pi_{0.5}$.

\begin{figure}[t]
    \centering
    \includegraphics[width=\linewidth]{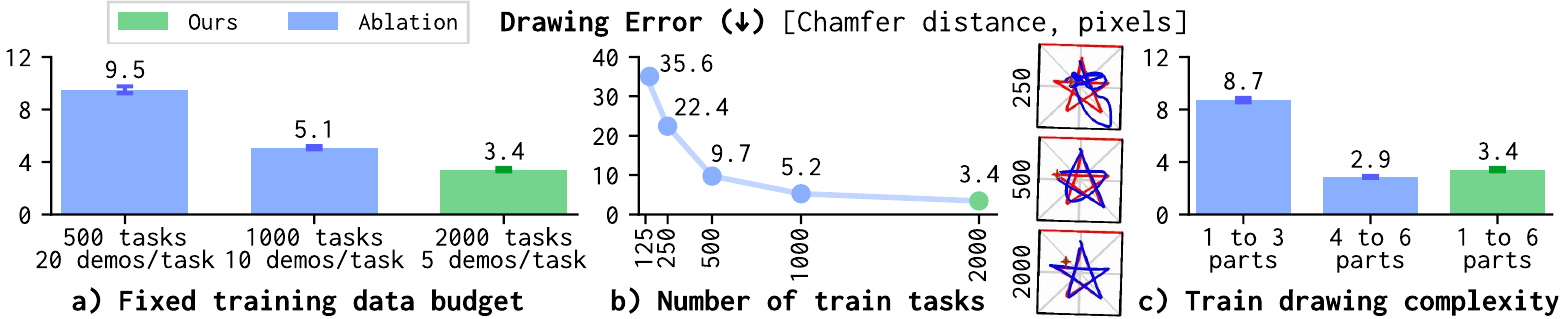}
    \vspace{-5mm}
    \caption{\textbf{Training data ablations on DrawAnything-Sim}. We report performance on \textit{unseen} drawings (3 seeds). Panel (b) uses 5 demos/task and shows representative qualitative examples.}
    \label{fig:data_ablations}
    \vspace{-2mm}
\end{figure}

\begin{figure}[t]
\begin{minipage}[c]{0.6\linewidth}
\centering
\resizebox{\linewidth}{!}{%
\begin{tabular}{llcc}
\hline
\textbf{Task} & \textbf{Result} & \textbf{Language} & \textbf{BPP (Ours)} \\
\hline
\multirow{4}{*}{\textit{Left arm}}
  & Success                                                                        & 24 (96\%)  & 19 (76\%) \\
  & Fail, dropped left sleeve                                                  &  1 (4\%)   &  4 (16\%) \\
  & Fail (a), did \textit{fold right arm}                                          &            &  1 (4\%)  \\
  & Fail (b), did \textit{fold bottom up}                                         &            &  1 (4\%)  \\
\hline
\textit{Right arm}
  & Success                                                                        & 25 (100\%) & 25 (100\%) \\
\hline
\multirow{4}{*}{\textit{Bottom up}}
  & Success                                                                        & 25 (100\%) & 15 (60\%)  \\
  & Fail (c), did \textit{fold right arm}                                          &            &  6 (24\%)  \\
  & Fail (d), right: grabs bottom \cmark,  &            & \multirow{2}{*}{4 (16\%)}  \\
  & \quad\quad\quad~ left: \textit{fold left arm} \xmark & & \\
\hline
\end{tabular}
}
\end{minipage}\hfill
\begin{minipage}[c]{0.37\linewidth}
    \centering
    \includegraphics[width=\linewidth]{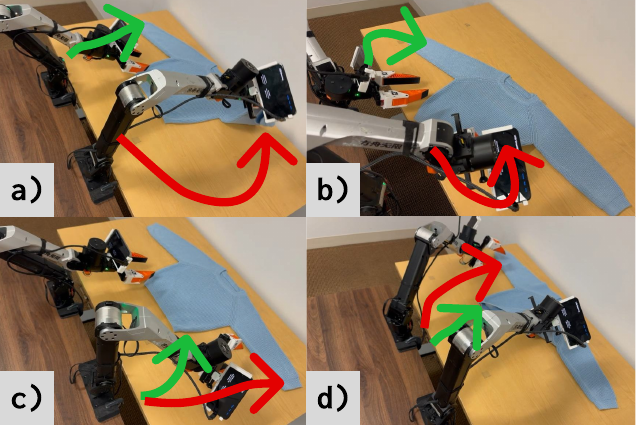}
\end{minipage}
\vspace{1mm}
\caption{\textbf{Laundry folding (low task diversity study).} Left: Results over 25 rollouts per task (one seed). Language conditioning succeeds on all tasks, while \ours occasionally competently completes the wrong fold; this implies weak task conditioning at low task diversity. Evaluations use \textit{unseen} prompts for \textit{seen} tasks. Right: corresponding failures; green is desired action, red is executed one.}
\vspace{-3mm}
\label{fig:laundry_eval}
\end{figure}

\textbf{\textit{Task diversity is more important than quantity per task.}} All ablations in this section use DrawAnything-Sim and may vary for other domains (for \liberogen, Fig.~\ref{fig:main_result}e gives a data ablation and Fig.~\ref{fig:attention_vis}b,d provides the attention analysis). \textsc{Data collection strategy.} For a fixed demonstration budget, collecting a few demonstrations for many drawing tasks improves test-time adaptation compared to many demonstrations of few tasks (Fig.~\ref{fig:data_ablations}a). We also find that increasing the number of training tasks improves performance on unseen tasks and that this adaptation to unseen tasks only emerges at high task diversity (Fig.~\ref{fig:data_ablations}b). These trends are consistent with task-diversity effects in sequence models~\citep{raventos2023pretraining} and in imitation learning~\citep{lin2025data}, and with concurrent industrial reports that attribute demonstration prompted in-context manipulation to task diversity~\citep{skild2026s1}. \textsc{Including complex training tasks.} We ablate the number of parts (line, circle, oval, Bézier curve) that are stitched together in the procedurally generated drawings that form the training set (Fig.~\ref{fig:data_ablations}c). We find that training only on low complexity drawings (1 to 3 parts) yields poor adaptation to complex, unseen drawings. Training on only complex drawings (4 to 6 parts) yields the best performance, but at the expense of a substantially larger demonstration dataset. The diversity required depends on the task space: for open-ended drawing, error keeps falling until roughly 1000 tasks (Fig.~\ref{fig:data_ablations}b), whereas \liberogen's discrete combinatorial tasks benefit from prompting with a few hundred. In either case, the ability to adapt to new tasks via prompting appears only above a domain-dependent level of task diversity.

\noindent\textbf{\textit{With low task diversity, task conditioning weakens.}}
\label{sec:laundry_folding}
\textsc{Laundry folding:} Behavior prompts are a far richer task representation than a fixed-length language embedding, so we ask whether that richness becomes a liability when task diversity is \textit{low}. We train on three sweater folding tasks collected with bimanual iPhUMI (Fig.~\ref{fig:teaser}): \textit{fold left arm}, \textit{fold right arm}, and \textit{fold bottom up}. All tasks start from a fully unfolded sweater, so the policy must identify the task from its descriptor alone. We compare against a language baseline which uses a CLIP encoding~\citep{radford2021learning}. Further task and evaluation details are in \S\ref{sec:appendix_laundry}.

\lang completes all three tasks at high success, while \ours succeeds on \textit{fold right arm} but sometimes executes the wrong fold on the other two (Fig.~\ref{fig:laundry_eval}). With only three tasks, the prompt's spatial and temporal detail does not further disambiguate the task; it adds variation in prompt duration and object configuration, which we postulate weakens task conditioning. The failures are in task identification, not motor competence: \ours performs the wrong fold competently. In \textit{fold bottom up}, for example, \ours grasps the bottom of the sweater correctly with the right arm, but the left arm then folds the left sleeve instead of also grasping the bottom, switching mid-execution to a different training task (Fig.~\ref{fig:laundry_eval}d). This result is expected given the task diversity ablations that show low task diversity is the hardest for prompting.

\subsection{How does the policy use the prompt?}
\label{sec:how}

\begin{figure*}[t]
    \centering
    \includegraphics[width=\linewidth]{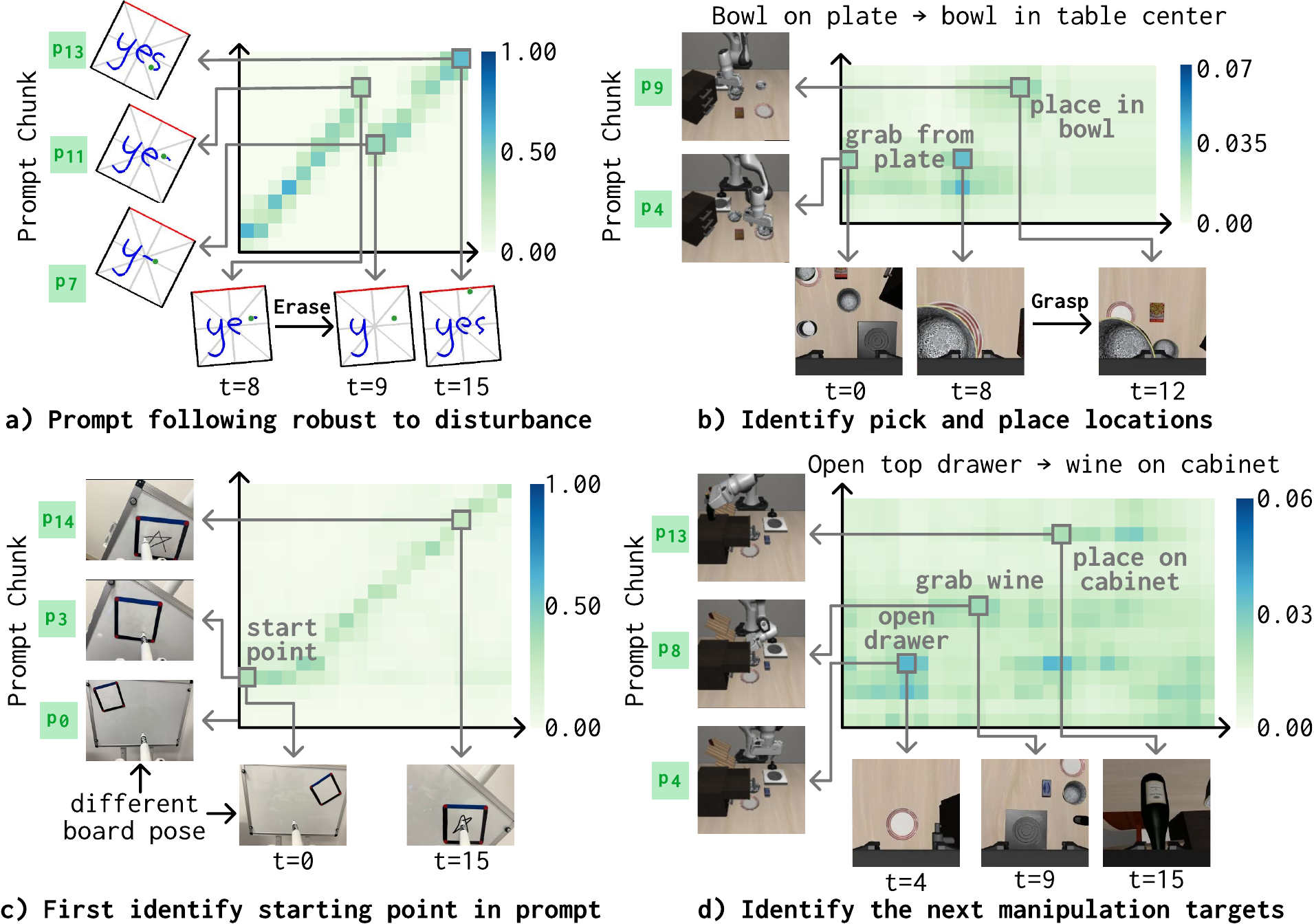} %
    \vspace{-3mm}
    \caption{\textbf{Prompt encoder attention (unseen tasks).} We visualize the normalized attention scores in the \ours prompt encoder on \textit{unseen} tasks during rollout (x-axis) to see how the model attends to the prompt (y-axis). For DrawAnything (a,c) the policy attention continuously tracks the portion of the prompt closest to the current observation, while in \liberogen (b: Combination, d: Chain) the attention tracks discrete task ``milestones''. The lower attention magnitude in \liberogen is due to learned attention sink tokens~\citep{xiao2024efficient} added to the start of the prompt (not shown).}
    \label{fig:attention_vis}
    \vspace{-2mm}
\end{figure*}

\noindent\textbf{\textit{BPP achieves test-time adaptation through dense sub-goal conditioning on the prompt.}}
In Fig.~\ref{fig:attention_vis}, we visualize the attention in the prompt encoder. For DrawAnything, the attention closely follows the task progression, indicating that \ours identifies \textit{temporal similarities} through a prompt lookup operation to find the section most similar to the current observation. As a result, the policy can extract upcoming states and actions that inform action generation after accounting for \textit{spatial differences} between the prompt and rollout board pose. This \textit{step-by-step guidance largely simplifies the learning complexity} compared to \goalimage, which must reconstruct complex tasks from just the final state. This lookup also explains the robustness to disturbance in \S\ref{sec:when}: when the canvas is erased mid-rollout (Fig.~\ref{fig:attention_vis}a, $t=8 \rightarrow 9$), the attention jumps back from prompt chunk $p_{11}$ to $p_7$, the chunk matching the reverted canvas, and then resumes advancing. This lookup is likely easier in \ours as the disturbance is not retained in the history, unlike \icrt. For \liberogen, we also find that the attention follows task progression, though in a more discrete manner to identify the next key event, such as a task transition, the next object to interact with, or where to place an object. These milestones are the steps a language command names, now grounded in the robot's observations.

\clearpage
\subsection{Why does the prompt help?}
\label{sec:why}

\begin{figure}[t]
    \centering
    \includegraphics[width=\linewidth]{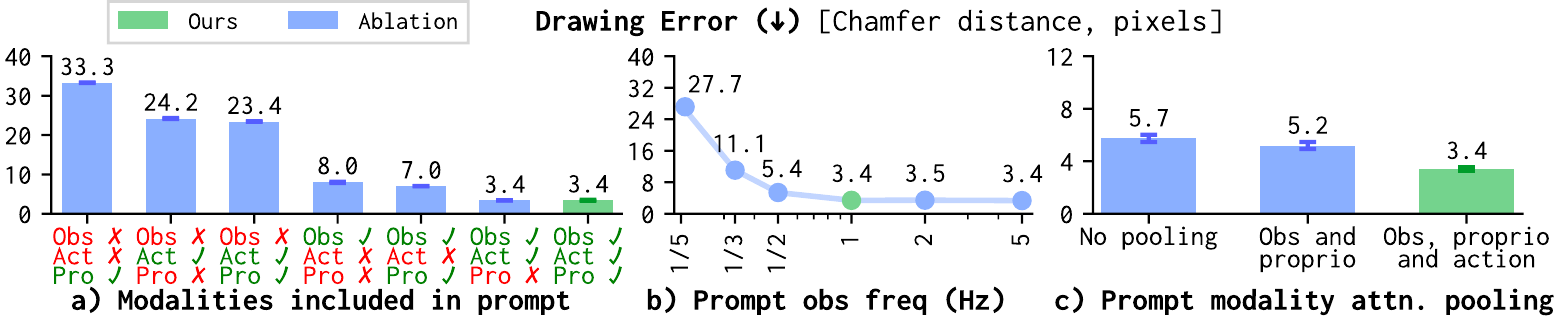}
    \vspace{-5mm}
    \caption{\textbf{Prompt ablations on DrawAnything-Sim}. We report performance on \textit{unseen} drawing tasks (3 seeds) while ablating what information is included in the prompt and the impact of applying attention pooling to the prompt tokens. We find that dense sensorimotor information in the prompt is critical for test-time adaptation and that pooling provides further improvement.}
    \label{fig:prompt_ablations}
    \vspace{-3mm}
\end{figure}

\noindent\textbf{\textit{The advantage comes from dense sensorimotor detail in the prompt.}} All ablations in this section use DrawAnything-Sim and results may vary for other domains. \textsc{Including multi-modal sensorimotor information} (Fig.~\ref{fig:prompt_ablations}a). For DrawAnything-Sim, we find that including observations in the prompt is necessary to anchor the prompt lookup and that including actions improves adaptation to unseen tasks. Proprioception (i.e., cursor position) is not useful as the cursor is already visually shown in the observation image. \textsc{Including observations at a sufficiently high frequency} (Fig.~\ref{fig:prompt_ablations}b). Aggressive downsampling of the prompt observations (less than 1Hz) makes \ours struggle to follow prompts for unseen drawings. \textsc{Using attention pooling to aggregate the modalities} (Fig.~\ref{fig:prompt_ablations}c). Attention pooling helps temporally associate modalities from the same prompt chunk and improves performance. It also reduces the sequence length by merging each prompt chunk into a single embedding, rather than having separate tokens per modality.

\textsc{Robustness to prompt quality:} Perturbing the prompt degrades performance only moderately: slowing the prompt by $1.25\times$ raises error from $3.43 \pm 0.10$ to $3.64 \pm 0.12$, adding trajectory jitter ($\sigma = 5$\,px) to $3.69 \pm 0.02$, and inserting three pauses to $4.26 \pm 0.12$, a 24\% increase (Tab.~\ref{tab:prompt_quality}).

\section{Limitations}
Our study has several limitations. First, in our benchmarks prompts and rollouts differ in object positions and board rotations but not in scene or object identity. A general-purpose behavior prompting policy trained on large amounts of diverse manipulation data would help study cases where the prompt is more different from the target environment. Second, we identify various levels of task diversity requirements across our two benchmarks, DrawAnything-Sim and \liberogen. Future work should identify what task properties determine the task diversity requirements for behavior prompting across many domains. Third, our attention analysis characterizes \ours; applying it to other in-context architectures would test whether dense sub-goal attention lookup is a general mechanism of behavior prompting. Fourth, unseen tasks in \liberogen recombine seen skills; a benchmark that isolates adaptation to entirely new tabletop action primitives would test whether a single prompt can teach one. Finally, in many cases, goal images or language may be sufficient to achieve the desired level of test-time adaptation and do not require collecting a full behavior prompt. Future studies should thus evaluate policies that can flexibly select among task descriptors at test time.

\section{Conclusion}

We presented an empirical study of behavior prompting. We find that task diversity, rather than demonstrations per task, is a key driver of prompting capability; that the policy uses the prompt as a source of dense sub-goals through an attention lookup over the prompt; and that dense sensorimotor detail in the prompt matters, with observations anchoring the lookup and actions improving fine-grained action adaptation. The task diversity finding reproduces claims in concurrent industrial reports on in-context demonstration prompting, while the sub-goal attention analysis and prompt composition ablations go beyond what these reports describe. All findings were obtained in reproducible settings without requiring industrial-scale data or compute. We release all benchmarks, data, and code so that the conditions enabling in-context behavior prompting can be further studied.

\newpage

\subsection*{Reproducibility statement}
We describe the \ours architecture, training procedure, and inference procedure in \S\ref{sec:prompting_policy}, and provide the training compute requirements and the prompt encoder and action decoder architecture specifications in Appendix~\S\ref{sec:appendix_training_details} (Tab.~\ref{tab:compute}, \ref{tab:pe_arch_appendix}, and \ref{tab:diffusion_arch_appendix}). Implementation details for all baselines (\goalimage, \lang, \icrt, and \pizerofive) are in Appendix~\S\ref{sec:appendix_baseline_details}, and the architectural comparison to \icrt is in Appendix~\S\ref{sec:bpp_comparison_to_prior_methods}. Our simulation benchmarks are procedurally generated: the DrawAnything-Sim training and evaluation tasks are shown in Appendix~\S\ref{sec:appendix_drawanything_sim}, and the \liberogen task generation, scripted demonstration generation, and the complete lists of held-out test tasks are in Appendix~\S\ref{sec:appendix_liberogen}. For the real-world experiments, the DrawAnything-Real data generation and evaluation procedure are in Appendix~\S\ref{sec:appendix_drawanything_real}, the laundry folding data and evaluation protocol are in \S\ref{sec:laundry_folding}, and the iPhUMI hardware, data collection, and deployment pipeline are in Appendix~\S\ref{sec:appendix_iphumi}. Simulation results are reported across three seeds (except where stated) with standard deviations (Fig.~\ref{fig:main_result} and \ref{fig:data_ablations} and~\ref{fig:prompt_ablations}). We have open-sourced all code and datasets, and all training hyperparameters are specified in the code release. The iPhUMI hardware and application software are open-sourced as well. All simulation results are directly reproducible with the open-sourced code and datasets.

\ificlrfinal
\subsubsection*{Acknowledgments}
We would like to thank the REAL lab for continuous project support, in particular Huy Ha for their insightful project guidance, Yihuai Gao for their help with the robot deployment setup, and Max Du for help with data collection. Regarding iPhUMI development, Yihuai Gao contributed USB streaming support and Xiaomeng Xu extended the bimanual data collection feature to support head-mounted iPhone data collection. We would also like to thank Chen Chen, Benoit Landry, Walter Talbott, and Jian Zhang for their feedback and discussions.
This work was supported in part by the
NSF Graduate Fellowship, NSF Award \#2143601, \#2037101, and \#2132519, and Apple. The views and conclusions contained herein are those of the authors and should not be interpreted as necessarily representing the official policies, either expressed or implied, of the sponsors.
\fi

\bibliography{references}
\bibliographystyle{iclr2027_conference}

\clearpage
\appendix
\begin{center}
{\Large\bfseries Appendix}
\end{center}

\section{DrawAnything-Real implementation}
\label{sec:appendix_drawanything_real}
\subsection{Scripted data generation}
For our real drawing task we train on a mixture of data collected via iPhUMI and a scripted policy. Specifically, we collect 200 tasks with 5 demonstrations per task from a human using iPhUMI. We also collect 800 tasks with 6 demonstrations per task using a scripted policy. The scripted policy maps procedurally generated drawings used in DrawAnything-Sim to open-loop commands in the 6DOF robot end-effector space. Using iPhUMI data during training enables the policy to work well with iPhUMI prompts coming from humans at test time and the scripted policy helps reach the task diversity required to enable adaptation to new tasks at test time.

For each scripted demonstration, a human operator places the drawing canvas at a random position and orientation on the whiteboard. We localize the four red corner markers on the drawing canvas with a segmentation model on an RGB image from the iPhone's main rear camera, yielding 2D keypoints in the camera frame that we use to estimate the canvas pose. We then transform the resulting pose into the robot base frame using a fixed camera-to-robot extrinsic calibration. The robot first moves to a random pose in front of the whiteboard, starts recording, then moves to the estimated canvas pose and aligns its end-effector orientation with the canvas. It executes the scripted drawing trajectory and finally retreats to a random pose. This is repeated across tasks and demonstrations.

\subsection{Evaluation procedure}
\begin{figure}[H]
    \centering
    \includegraphics[width=\linewidth]{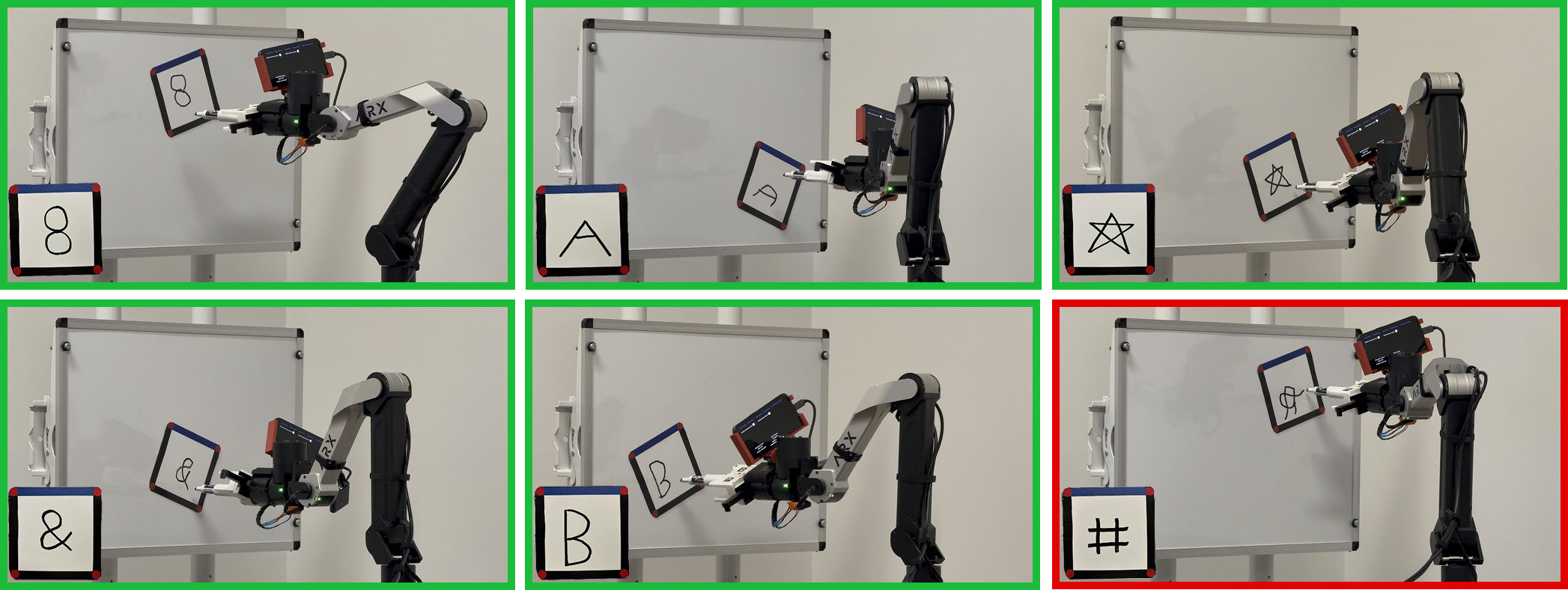}
    \vspace{-5mm}
    \caption{\textbf{\ours qualitative results on unseen tasks in DrawAnything-Real.} We find that \ours is able to reconstruct unseen drawings given a single iPhUMI demo. We show successful drawings (green) and a failure case (red).}
    \label{fig:draw_real_unseen}
\end{figure}

\begin{figure}[H]
    \centering
    \includegraphics[width=0.9\linewidth]{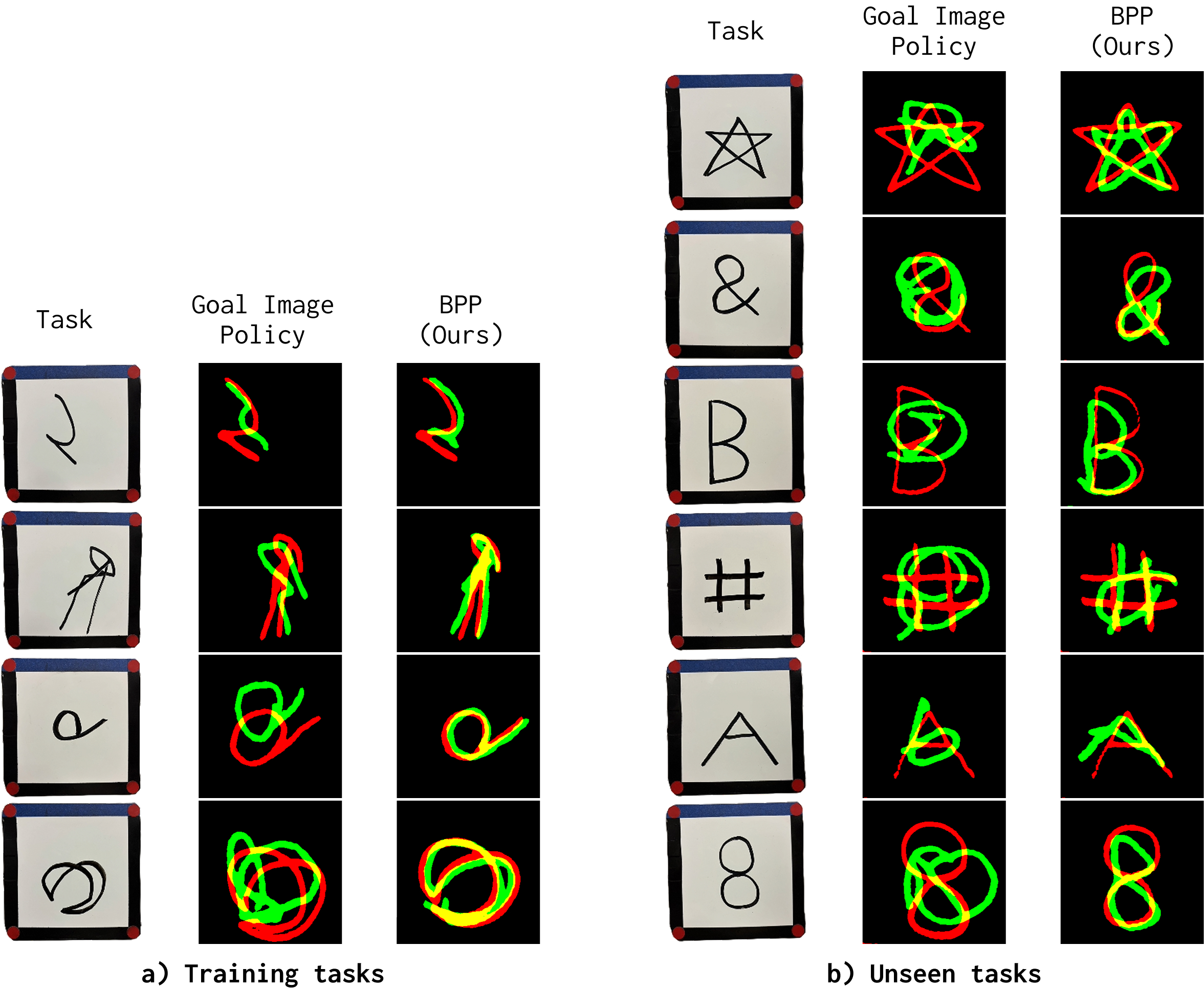}
    \caption{\textbf{DrawAnything-Real evaluation tasks with representative examples of robot executions.} Green: executed drawing by policy. Red: goal image input rendered as a reference overlay. We observe that \goalimage can roughly match the structure of the training drawings, but fails to replicate unseen drawings. BPP is able to reconstruct both training and unseen drawings given a single demonstration.}
    \label{fig:qual_goal_vs_prompt_train}
\end{figure}

Qualitative results of \ours on some unseen drawings are shown in Fig.~\ref{fig:draw_real_unseen}. Training and test tasks used for evaluation are shown in Fig.~\ref{fig:qual_goal_vs_prompt_train}. For each evaluation trial, the drawing canvas is placed at a random position and orientation on the whiteboard. We sample a single demonstration to define a matched conditioning pair: the full demonstration is used as the behavior prompt for the  prompting policy, and the corresponding goal image extracted from that same demonstration is used for the goal-image policy.

To control for initial conditions, we execute the prompting policy, reset the robot and environment to the same initial state, and then execute the goal-image policy with the matched goal image. We repeat this procedure after moving the canvas to a new random placement and sampling a new prompt/goal-image pair.

For evaluation on training tasks, conditioning pairs are sampled from the set of six demonstrations collected from four of the procedural training tasks. For evaluation on unseen test tasks, conditioning pairs are sampled from a set of three iPhUMI demonstrations collected for each of the six unseen evaluation tasks. Ten evaluation rollouts are done for each task.

\clearpage

\subsection{Qualitative comparison: goal image vs. behavior prompting}
Fig.~\ref{fig:qual_goal_vs_prompt_train} shows representative examples comparing \ours to \goalimage on training and unseen drawing tasks. In DrawAnything-Sim, we observed that \goalimage performs well on training tasks (Fig.~\ref{fig:main_result}a). In DrawAnything-Real, we find however that \goalimage underperforms quantitatively on training tasks (Fig.~\ref{fig:main_result}b). However, qualitative inspection of Fig.~\ref{fig:qual_goal_vs_prompt_train}a suggests a consistent failure mode: the policy often produces a recognizable rendition (or partial rendition) of the intended shape, but with spatial misalignment or local geometric errors that cause large Chamfer distances.

For unseen tasks, \goalimage frequently degenerates into near-random strokes that do not resemble the target shape, whereas \ours produces coherent drawings aligned with the demonstrated behavior. Interestingly, we find that \goalimage has similar quantitative error rates between training and unseen tasks (Fig.~\ref{fig:main_result}b). However, we see that unlike on training tasks, \goalimage has poor qualitative fidelity on these test tasks (Fig.~\ref{fig:qual_goal_vs_prompt_train}b). We believe this is because the unintelligible drawings produced by \goalimage on test tasks can achieve moderate levels of Chamfer error by roughly covering the target drawing region without reproducing the intended structure.

\clearpage

\section{DrawAnything-Sim Tasks}
\label{sec:appendix_drawanything_sim}
\textbf{Metric.} We report the Chamfer distance in pixels between the drawn canvas and the target drawing, taking the lowest distance reached during the rollout, averaged over 32 environment seeds and 50 unseen tasks.

\begin{figure}[H]
    \centering
    \includegraphics[width=\linewidth]{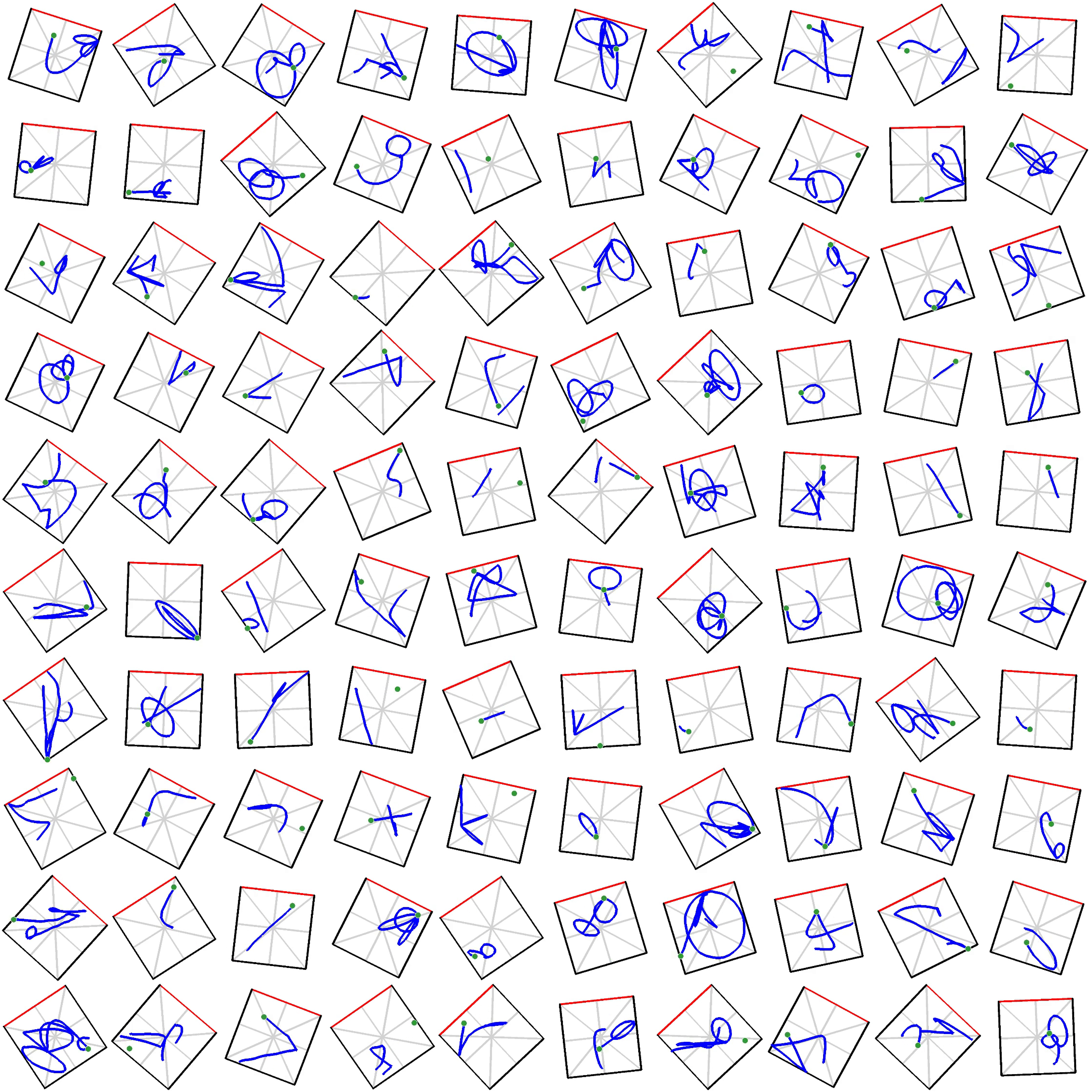}
    \caption{\textbf{DrawAnything-Sim training tasks.} We show a subset of 100 of the 2000 procedurally generated tasks (each having 5 demonstrations per task) using a combination of 1 to 6 parts. Parts include lines, Bézier curves, partial/full ovals, and free space (pen up) movement. The parameters for each part (such as start/end position, Bézier control points, proportion of oval, clockwise/counter-clockwise direction, etc.) are randomly sampled. With a specified probability, parts will connect back to the start/end position of a previous part. Each demonstration includes varied whiteboard rotation (from $-\frac{\pi}{4} to \frac{\pi}{4}$), has varied speed, and includes noise inserted into the trajectories.}
    \label{fig:DrawAnything_sim_train_dataset}
\end{figure}

\begin{figure}[H]
    \centering
    \includegraphics[width=0.8\linewidth]{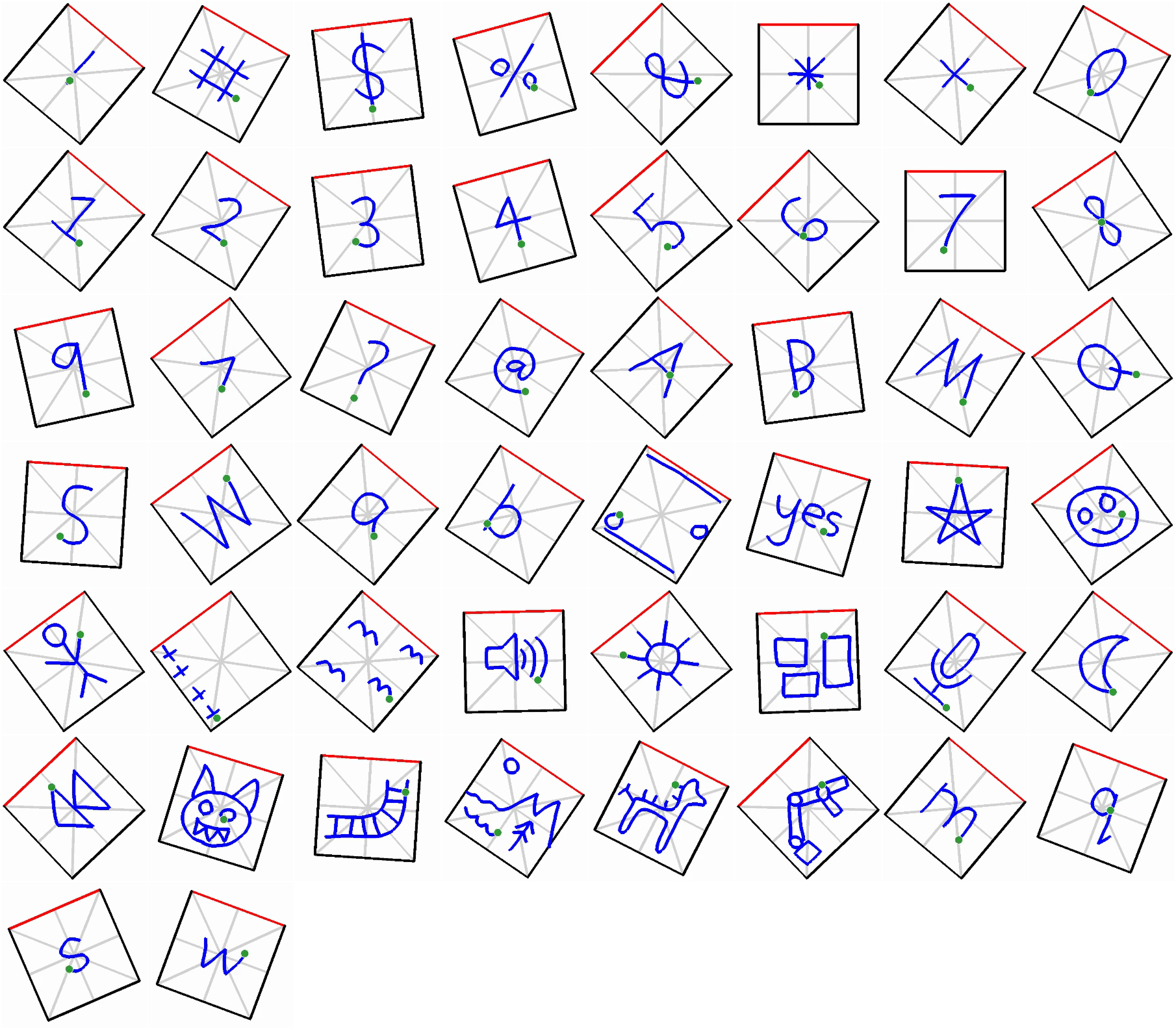}
    \caption{\textbf{DrawAnything-Sim evaluation tasks.} We hand-collect 50 evaluation tasks that were not seen during training with 5 demonstrations per task at varying board rotations. Tasks have varied complexity and duration.}
    \label{fig:DrawAnything_sim_eval_dataset}
\end{figure}

We visualize the training (Fig.~\ref{fig:DrawAnything_sim_train_dataset}) and evaluation dataset (Fig.~\ref{fig:DrawAnything_sim_eval_dataset}) for DrawAnything-Sim.

\clearpage

\clearpage

\section{Results on original LIBERO dataset}
We report results on the original LIBERO~\citep{liu2023libero} dataset. The original LIBERO tasks are sufficiently described by language and the original benchmark splits do not evaluate test-time adaptation to unseen tasks. As expected, we find that using behavior prompting achieves similar performance to the language-conditioned diffusion policy. We empirically found that finetuning a CLIP language encoder gives substantially better performance compared to the frozen language embeddings used in some prior diffusion policy LIBERO evaluations~\citep{kim24openvla}. Modern policies have also started to saturate performance of the original LIBERO benchmark~\citep{black2026pi0visionlanguageactionflowmodel,kim2025fine}, emphasizing the importance of LIBERO-Gen which extends the dataset to evaluate test-time adaptation to unseen tasks.

\begin{table}[H]
\centering
\begin{tabular}{lcc|c}
\hline
\textbf{Split} & \textbf{Language Diffusion Policy} & \textbf{Behavior Prompting Policy (Ours)} & \textbf{$\bm{\pi_{0.5}}$} \\ \hline
LIBERO Spatial & $97.80 \pm 0.71$ & $97.40 \pm 1.02$ & $97.8$ \\
LIBERO Object  & $97.80 \pm 0.49$ & $98.93 \pm 0.52$ & $97.4$ \\
LIBERO Goal    & $95.60 \pm 1.07$ & $98.33 \pm 0.38$ & $97.8$ \\
LIBERO 10      & $93.73 \pm 2.07$ & $95.27 \pm 0.66$ & $95.6$ \\ \hline
Average        & $96.23 \pm 0.55$ & $97.48 \pm 0.33$ & $97.15$ \\ \hline
\end{tabular}
\caption{\textbf{Original LIBERO dataset results.} Model success rate (\%) on the original LIBERO~\citep{liu2023libero} dataset by suite across 3 seeds (mean $\pm$ stdev). \lang is detailed in~\S\ref{sec:appendix_baseline_details}. We finetuned $\pi_{0.5}$~\citep{intelligence2025pi05visionlanguageactionmodelopenworld} for 30K LoRA~\citep{hu2022lora} steps using openpi and report results for one seed. The $\pi_{0.5}$ training does not include LIBERO-90 in the finetuning mix, while the other models presented include LIBERO-90 split in the training data. Due to this difference, we put $\pi_{0.5}$ results in a separated column. For each of the three policy architectures listed we train one checkpoint per seed across all of the listed LIBERO splits rather than training and evaluating separate checkpoints per split.}
\label{tab:libero_results_std}
\end{table}

\section{\liberogen implementation details}
\label{sec:appendix_liberogen}
\liberogen procedurally generates new task definitions consisting of an environment definition and execution steps (\S\ref{sec:liberogen_procedural_task_gen}). It also generates corresponding demonstrations for those tasks (\S\ref{sec:liberogen_procedural_demo_gen}). We then discuss details for \liberogencombination (\S\ref{sec:liberogen_combination}) and \liberogenchain (\S\ref{sec:liberogen_chain}), two benchmarks created using this tool.

\subsection{Procedural task generation}
\label{sec:liberogen_procedural_task_gen}

The 130 original tasks in the LIBERO dataset are defined using BDDL, which defines both the environment and the intended robot task. The environment is defined by a base scene, objects present in that scene, and the initial locations of those objects. The task is defined through a set of predicates that specify the desired final states of the objects in the scene. \liberogen takes these BDDL files and enumerates the possible initial object locations and the desired final object states to generate new task definitions. The user specifies the set of procedural operations they would like to occur to generate a desired set of task variation, such as: 1) \textit{generate environments where the goal is the same, but the plate is initialized at all of the possible starting locations} or 2) \textit{generate environments where the initial object configuration stays the same, but the goal states are all the places the cream cheese could be placed}. \liberogen also supports the generation of chained tasks where the goal states consist of multiple action primitives applied sequentially (e.g., \textit{turn on the stove and then put the bowl on the stove}).

In practice, we manually define a set of valid operators for each object (e.g., bowls can be grasped or you can put an object in them). We also start with a base set of environments from existing LIBERO tasks for which we apply the variations on top of. Doing so typically creates a large combinatorial space of tasks which either 1) may extend beyond the desired set of tasks for a targeted scientific experiment or 2) may want to be partitioned into a training and test set. We address this through the use of \textit{views} which include filters that select desired portions of the generated tasks to put into training or testing splits.

\subsection{Procedural demonstration generation}
\label{sec:liberogen_procedural_demo_gen}

Given the procedurally generated task definitions, we need to procedurally generate demonstration data for these tasks. We do so through a scripted policy that leverages ground simulator object states and the existing demonstration data for the 130 original LIBERO tasks. From the existing teleoperation data, we extract object-relative grasp poses for each of the target objects. Many tasks also involve replaying a portion of teleoperation data and then switching to a scripted policy to complete the new part of the task. We introduce variations at multiple points to simulate differences present within teleoperated demonstrations: grasp location, placement position, intermediate poses on the way to or from each action primitive, and initial robot and object states. We validate the success of each demonstration using the predicate checking mechanism present in LIBERO. For chained tasks, the BDDL does not encode the ordering of each action primitive, so we encode the desired sequential order in a separate configuration file.

\subsection{\liberogencombination test tasks}
\label{sec:liberogen_combination}

\begin{figure}[H]
    \centering
    \includegraphics[width=\linewidth]{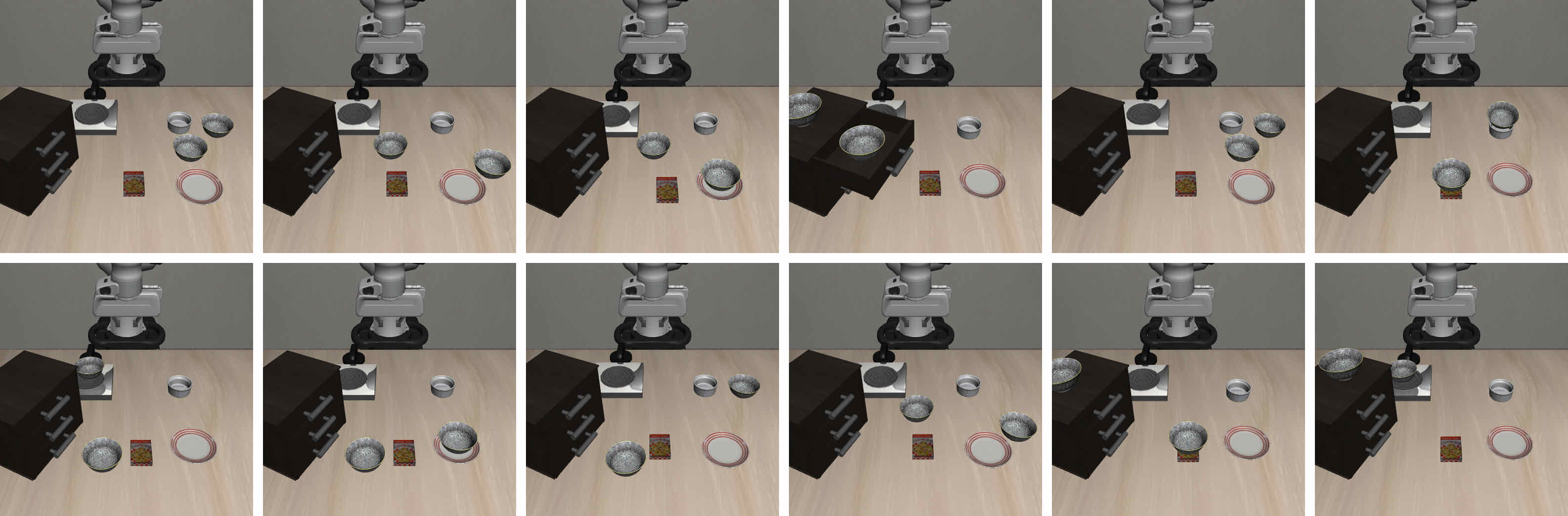}
    \vspace{-6mm}
    \caption{\textbf{Initial environments for \liberogencombination.} There are two identical bowls in each environment. The task involves moving a specified bowl to a specified target location.}
    \label{fig:liberogen_combination_inits}
\end{figure}

Here we extend the discussion of \liberogencombination from \S\ref{sec:evaluation}. We modify the LIBERO Spatial environments to have additional starting bowl locations as pictured in Fig.~\ref{fig:liberogen_combination_inits}.

From each starting initialization, we generate the combinatorial space of all possible bowl pick locations and place locations to generate a task distribution. We withhold ten tasks (combinations of pick-place locations) to use as evaluation tasks. We document each of these unseen tasks below including the count of training tasks that have the same pick location and training tasks that have the same place location; this indicates that the pick and place locations have individually been seen during training, but never jointly. This experiment design evaluates the capability of the policy to adapt to unseen instructions at test time without evaluating unseen action primitives.\\

\textbf{Unseen training tasks for \liberogencombination and relevant training tasks:}

\begin{Verbatim}[fontsize=\scriptsize, breaklines=true]
Unseen: Pick the bowl from table center and place it on the cookies box
  Training: Same pick (bowl table center), different place [9 tasks]
  Training: Different pick, same place (bowl on the cookies box) [8 tasks]

Unseen: Pick the bowl in the top layer of the wooden cabinet and place it on the plate
  Training: Same pick (bowl wooden cabinet top), different place [7 tasks]
  Training: Different pick, same place (bowl on the plate) [9 tasks]

Unseen: Pick the bowl next to the cookies box and place it on the bowl next to the ramekin
  Training: Same pick (bowl next to box), different place [9 tasks]
  Training: Different pick, same place (bowl on the bowl next to the ramekin) [4 tasks]

Unseen: Pick the bowl next to the plate and place it on the stove
  Training: Same pick (bowl next to plate), different place [9 tasks]
  Training: Different pick, same place (bowl on the stove) [9 tasks]

Unseen: Pick the bowl on the cookies box and place it on the bowl on the wooden cabinet
  Training: Same pick (bowl cookies), different place [9 tasks]
  Training: Different pick, same place (bowl on the bowl on the wooden cabinet) [8 tasks]

Unseen: Pick the bowl on the cookies box and place it on the wooden cabinet
  Training: Same pick (bowl cookies), different place [9 tasks]
  Training: Different pick, same place (bowl on the wooden cabinet) [8 tasks]

Unseen: Pick the bowl on the plate and place it on the bowl on the table center
  Training: Same pick (bowl plate), different place [9 tasks]
  Training: Different pick, same place (bowl on the bowl on the table center) [4 tasks]

Unseen: Pick the bowl on the plate and place it on the table center
  Training: Same pick (bowl plate), different place [9 tasks]
  Training: Different pick, same place (bowl on the table center) [7 tasks]

Unseen: Pick the bowl on the stove and place it on the ramekin
  Training: Same pick (bowl stove), different place [9 tasks]
  Training: Different pick, same place (bowl on the ramekin) [9 tasks]

Unseen: Pick the bowl on the wooden cabinet and place it between the plate and the ramekin
  Training: Same pick (bowl wooden cabinet top side), different place [8 tasks]
  Training: Different pick, same place (bowl between the plate and the ramekin) [9 tasks]
\end{Verbatim}

\subsection{\liberogenchain test tasks}
\label{sec:liberogen_chain}

\begin{figure}[H]
    \centering
    \includegraphics[width=\linewidth]{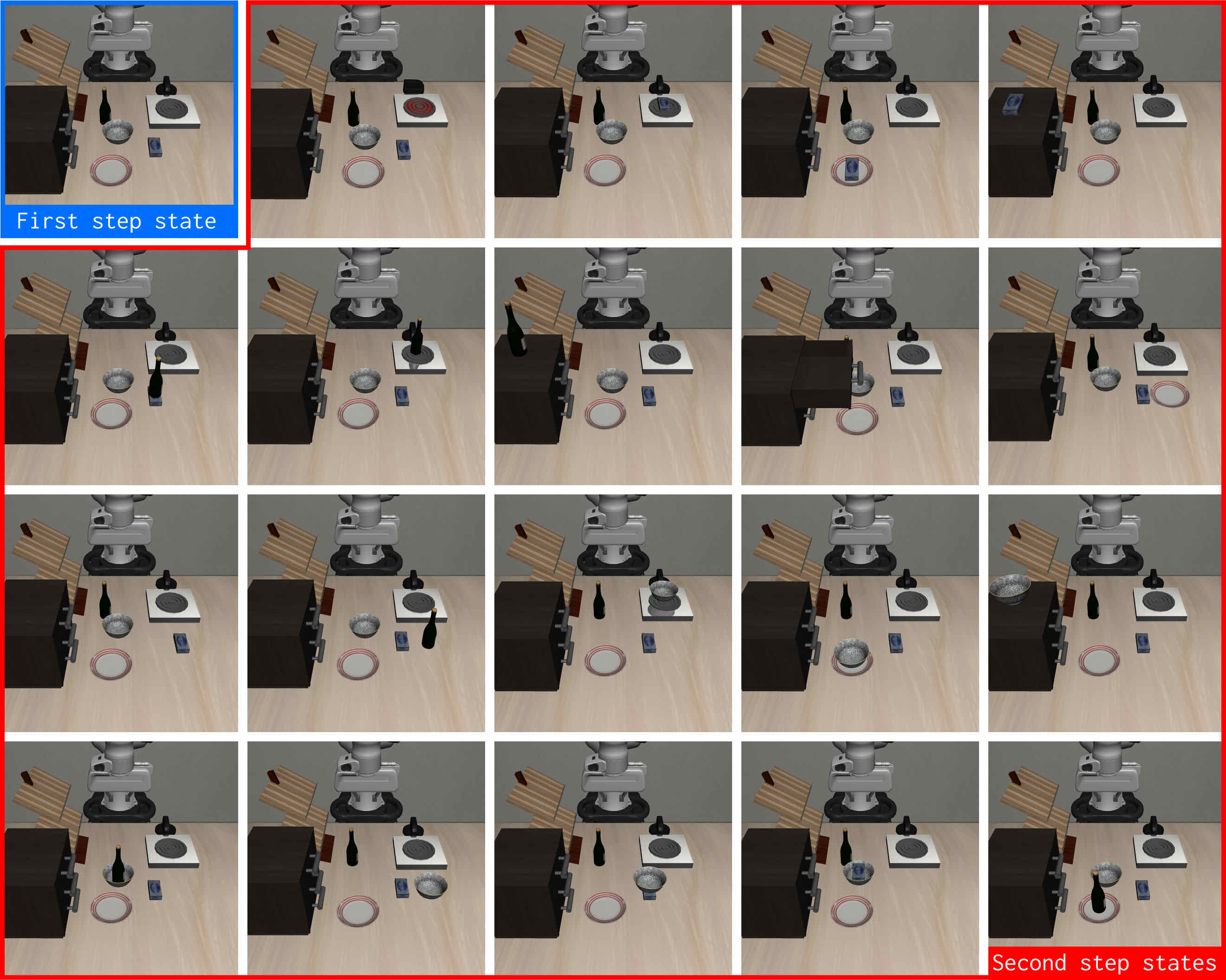}
    \vspace{-5mm}
    \caption{\textbf{Initial environment for \liberogenchain.} All two-step chained tasks in this experiment start from the single first step state (blue). We also include one step actions starting from the first step state (blue) as well as second step actions which start assuming one action primitive has already been completed (red). The ablation for no second step excludes the single-step tasks that start from the second step initializations (red).}
    \label{fig:liberogen_chain_inits}
\end{figure}

Here we extend the discussion of \liberogenchain from \S\ref{sec:evaluation}. In \liberogenchain we have one environment initialization and enumerate the set of two-step sequences where we chain together two action primitives sequentially (see Fig.~\ref{fig:liberogen_chain_inits}).

At test time, we evaluate adaptation to \textit{unseen} sequential chains of individually \textit{known} primitives. The purpose is not to adapt to an entirely new action primitive at test time that has not been seen during training, but rather to adapt to a new sequence of these primitives. This setup evaluates the capability of a model to follow previously unseen instructions. Here we enumerate all ten unseen evaluation tasks and, for each one, include the counts of the similar tasks that the model has seen during training. In short, this list demonstrates that the primitives for each unseen task have been seen individually during training, but never sequentially chained. Our ablation where we remove the second step tasks from the training mixture removes the guarantee that the second step task in the chain has been seen with the exact same environment initialization (see Fig.~\ref{fig:liberogen_chain_inits}).

\textbf{Unseen training tasks for \liberogenchain and relevant training tasks:}

\begin{Verbatim}[fontsize=\scriptsize, breaklines=true]
Unseen: [1st] Open the middle layer of the drawer and then [2nd] put the cream cheese on the stove
  Training: similar chain (same 1st, different 2nd) [3 tasks]:
  -  Open the middle layer of the drawer and put the cream cheese inside
  -  Open the middle layer of the drawer and then put the cream cheese in front of the stove
  -  Open the middle layer of the drawer and then put the cream cheese on the top of the drawer
  Training: similar single-step (2nd step match with wooden cabinet middle open): not present
  Training: similar single-step (2nd step match with diff background object config): present

Unseen: [1st] Open the top layer of the drawer and then [2nd] put the wine bottle on the top of the drawer
  Training: similar chain (same 1st, different 2nd) [5 tasks]:
  -  Open the top layer of the drawer and put the wine bottle inside
  -  Open the top layer of the drawer and then put the wine bottle in front of the stove
  -  Open the top layer of the drawer and then put the wine bottle on the bowl
  -  Open the top layer of the drawer and then put the wine bottle on the cream cheese
  -  Open the top layer of the drawer and then put the wine bottle on the stove
  Training: similar single-step (2nd step match with wooden cabinet top open): present
  Training: similar single-step (2nd step match with diff background object config): present

Unseen: [1st] Push the plate to the front of the stove and then [2nd] put the bowl on the plate
  Training: similar chain (same 1st, different 2nd) [3 tasks]:
  -  Push the plate to the front of the stove and then put the bowl on the cream cheese
  -  Push the plate to the front of the stove and then put the bowl on the stove
  -  Push the plate to the front of the stove and then put the bowl on the top of the drawer
  Training: similar single-step (2nd step match with plate on stove front): present
  Training: similar single-step (2nd step match with diff background object config): present

Unseen: [1st] Put the bowl on the plate and then [2nd] put the cream cheese on the bowl
  Training: similar chain (same 1st, different 2nd) [4 tasks]:
  -  Put the bowl on the plate and then put the cream cheese in front of the stove
  -  Put the bowl on the plate and then put the cream cheese on the bowl region
  -  Put the bowl on the plate and then put the cream cheese on the stove
  -  Put the bowl on the plate and then put the cream cheese on the top of the drawer
  Training: similar single-step (2nd step match with bowl on plate): present
  Training: similar single-step (2nd step match with diff background object config): present

Unseen: [1st] Put the bowl on the top of the drawer and then [2nd] put the cream cheese on the plate
  Training: similar chain (same 1st, different 2nd) [4 tasks]:
  -  Put the bowl on the top of the drawer and then put the cream cheese in front of the stove
  -  Put the bowl on the top of the drawer and then put the cream cheese on the bowl
  -  Put the bowl on the top of the drawer and then put the cream cheese on the bowl region
  -  Put the bowl on the top of the drawer and then put the cream cheese on the stove
  Training: similar single-step (2nd step match with bowl on wooden cabinet top side): present
  Training: similar single-step (2nd step match with diff background object config): present

Unseen: [1st] Put the cream cheese on the plate and then [2nd] put the bowl on the cream cheese
  Training: similar chain (same 1st, different 2nd) [4 tasks]:
  -  Put the cream cheese on the plate and then put the bowl in front of the stove
  -  Put the cream cheese on the plate and then put the bowl on the cream cheese region
  -  Put the cream cheese on the plate and then put the bowl on the stove
  -  Put the cream cheese on the plate and then put the bowl on the top of the drawer
  Training: similar single-step (2nd step match with cream cheese on plate): present
  Training: similar single-step (2nd step match with diff background object config): present

Unseen: [1st] Put the cream cheese on the top of the drawer and then [2nd] put the wine bottle on the bowl
  Training: similar chain (same 1st, different 2nd) [5 tasks]:
  -  Put the cream cheese on the top of the drawer and then put the wine bottle in front of the stove
  -  Put the cream cheese on the top of the drawer and then put the wine bottle on the cream cheese
  -  Put the cream cheese on the top of the drawer and then put the wine bottle on the cream cheese region
  -  Put the cream cheese on the top of the drawer and then put the wine bottle on the plate
  -  Put the cream cheese on the top of the drawer and then put the wine bottle on the stove
  Training: similar single-step (2nd step match with cream cheese on wooden cabinet top side): present
  Training: similar single-step (2nd step match with diff background object config): present

Unseen: [1st] Put the wine bottle in front of the stove and then [2nd] put the bowl on the stove
  Training: similar chain (same 1st, different 2nd) [3 tasks]:
  -  Put the wine bottle in front of the stove and then put the bowl on the plate
  -  Put the wine bottle in front of the stove and then put the bowl on the wine bottle region
  -  Put the wine bottle in front of the stove and then put the bowl on the top of the drawer
  Training: similar single-step (2nd step match with wine bottle on stove front): present
  Training: similar single-step (2nd step match with diff background object config): present

Unseen: [1st] Put the wine bottle on the rack and then [2nd] put the bowl on the top of the drawer
  Training: similar chain (same 1st, different 2nd) [5 tasks]:
  -  Put the wine bottle on the rack and then put the bowl in front of the stove
  -  Put the wine bottle on the rack and then put the bowl on the cream cheese
  -  Put the wine bottle on the rack and then put the bowl on the plate
  -  Put the wine bottle on the rack and then put the bowl on the stove
  -  Put the wine bottle on the rack and then put the bowl on the wine bottle region
  Training: similar single-step (2nd step match with wine bottle on wine rack top): not present
  Training: similar single-step (2nd step match with diff background object config): present

Unseen: [1st] Turn on the stove and then [2nd] put the bowl on the stove
  Training: similar chain (same 1st, different 2nd) [4 tasks]:
  -  Turn on the stove and then put the bowl in front of the stove
  -  Turn on the stove and then put the bowl on the cream cheese
  -  Turn on the stove and then put the bowl on the plate
  -  Turn on the stove and then put the bowl on the top of the drawer
  Training: similar single-step (2nd step match with flat stove turned on): present
  Training: similar single-step (2nd step match with diff background object config): present
\end{Verbatim}

\clearpage

\section{Baseline details}
\label{sec:appendix_baseline_details}
Our \lang and \goalimage baseline models follow the same diffusion CNN U-Net architecture from~\citet{chi2023diffusionpolicy} which also matches the architecture of the \ours action decoder. In other words, \ours is a standard diffusion policy with a prompt encoder attached, and the same action decoder serves all three task descriptors, enabling controlled comparisons across them. The difference between \ours and the baselines is the observation encoding; in place of a prompt encoder, \lang uses a finetuned CLIP~\citep{radford2021learning} language encoder and \goalimage shares a finetuned CLIP vision encoder with the corresponding image input.

LIBERO:
\begin{itemize}
    \item \lang uses a fine-tuned CLIP encoding which we found performed substantially better than frozen language embeddings. For specific LIBERO splits where it's not possible to use the initial environment state to infer the task (such as LIBERO Goal), we observed that freezing the language encoding would often cause the policy to complete the wrong task.
    \item \goalimage shares the same fine-tuned vision encoder used by the 3rd person camera. The goal image is the final frame of a demonstration from the 3rd person camera. Using goal images has ambiguity for some tasks within the original LIBERO dataset, so we only provide results for this baseline for \liberogencombination and \liberogenchain which do not contain tasks for which the goal image is ambiguous about the final state. For \liberogenchain, goal images have ambiguity as they do not indicate the ordering in which the two chained operations should occur.
    \item \pizerofive is finetuned using LoRA~\citep{hu2022lora}. For training \pizerofive we follow the LIBERO preprocessing steps from~\citep{kim24openvla} to regenerate our \liberogen datasets at a higher resolution, but do not filter out no-op operations. We finetune for 100K steps for \liberogen experiments (Fig.~\ref{fig:main_result}) and finetune for 30K steps for the results on the original LIBERO dataset (Tab.~\ref{tab:libero_results_std}).
\end{itemize}

DrawAnything-Sim:
\begin{itemize}
    \item \goalimage shares the fine-tuned vision encoder with the main image encoder.
    \item \icrt extends the provided implementation by~\citet{fu2024icrt} to support action chunking at the token input level; this means that 1) each input action token contains N sequential action steps rather than a single step and 2) we no longer encode observation tokens after every step, only after every N action steps. Additionally, at test time we only provide a single, full demonstration in the context at test-time. Both changes enable a more direct comparison to \ours, which chunks actions together in the prompt and uses a single behavior prompt during deployment.
\end{itemize}

DrawAnything-Real:
\begin{itemize}
    \item \goalimage shares the vision encoder with the iPhone main RGB camera in real. The drawing is not fully visible in all frames of the demonstrations due to the moving, wrist-mount camera and gripper occlusion. Thus, to ensure fair comparison, we select the goal image as the first frame searching backwards from the end of demonstration that fully shows the drawing region.
\end{itemize}

Laundry Folding:
\begin{itemize}
    \item \lang uses a finetuned CLIP language encoding.
\end{itemize}

\clearpage

\section{Prompt-quality sensitivity}
\label{sec:appendix_prompt_quality}
We perturb the behavior prompt at test time on DrawAnything-Sim (50 unseen tasks, 3 seeds) and report the drawing error of \ours (Tab.~\ref{tab:prompt_quality}). Slowing the prompt, jittering its trajectory, and inserting pauses each degrade performance only moderately relative to the clean prompt.
\begin{table}[H]
\centering
\begin{tabular}{lcc}
\hline
\textbf{Prompt perturbation} & \textbf{Error ($\downarrow$, mean $\pm$ std over 3 seeds)} & \textbf{$\Delta$ vs.~clean} \\
\hline
Clean prompt & $3.43 \pm 0.10$ & -- \\
$1.25\times$ slowdown & $3.64 \pm 0.12$ & $+5.9\%$ \\
Trajectory jitter ($\sigma = 5$\,px) & $3.69 \pm 0.02$ & $+7.4\%$ \\
Three inserted pauses (10 steps each) & $4.26 \pm 0.12$ & $+24.0\%$ \\
\hline
\end{tabular}
\vspace{1mm}
\caption{\textbf{Prompt-quality sensitivity on DrawAnything-Sim.} Drawing error on 50 unseen tasks (3 seeds) when the test-time prompt is perturbed.}
\label{tab:prompt_quality}
\end{table}

\section{BPP architectural comparison to ICRT}
\label{sec:bpp_comparison_to_prior_methods}

\textbf{Model architecture.} ICRT~\citep{fu2024icrt} is a prior behavior prompting architecture that we baseline against. ICRT is a causal transformer decoder, while BPP has separate modules for prompt understanding and action generation (detailed in \S\ref{sec:prompting_policy}). \ours leverages cross-attention to reason over the prompt contents while ICRT leverages causal self-attention. Furthermore, BPP leverages action diffusion~\citep{chi2023diffusionpolicy}, while ICRT uses L1 action loss. 

\textbf{Training.} Each step of training, ICRT samples a sequence of prompts and entire rollouts concatenated along the time dimension into a single sequence to fill the transformer context length. The loss is computed by predicting action loss for each step of the rollout sequence using teacher forcing. In contrast, for each training step of \ours, we sample a single behavior prompt along with a batch of observation-action pairs randomly sampled across many demonstrations. Action loss is computed for each of these randomly sampled observation-action pairs. Both models use prompts and rollouts from a single task per batch. In practice, due to DDP training across multiple GPUs or the use of gradient accumulation, each gradient update will take into account losses across multiple tasks.

\textbf{Inference.} \ours has a fixed length history, common in other visuomotor policies~\citep{chi2023diffusionpolicy}, whereas ICRT retains the full rollout history in the model context. As a result, ICRT's maximum execution duration is tied to its training-time context length, while \ours can be rolled out indefinitely. \S\ref{sec:when} reports the behavior of both models under a mid-rollout disturbance; the protocol is given at the end of this section.

\ours selects a single prompt per rollout, encodes it once, and caches the chunk embeddings; the prompt encoder then extracts relevant prompt information once per inference call, after which action denoising proceeds without referencing the prompt. Each forward pass of ICRT attends over the prompt and the full rollout history, with KV caching. Per inference call, ICRT is faster (6.5\,ms with a single L1 regression pass versus 49\,ms for \ours, which uses 16 denoising steps), and ICRT's per-call cost stays flat over a rollout thanks to KV caching at our scale. We therefore make no efficiency claim for \ours; the practical difference is the unbounded rollout duration and the use of action diffusion.

Full-history policies can also overfit to spurious correlations in training rollouts~\citep{torne2025learninglongcontextdiffusionpolicies}. We find little evidence of this here: replacing \ours's diffusion head with L1 regression accounts for most of the gap to ICRT on unseen DrawAnything-Sim drawings (error of 4.84 for \ours with an L1 head vs.\ 5.07 for ICRT, compared to 3.43 for \ours), leaving little to attribute to history.

\textbf{Disturbance protocol.} On the 50 unseen DrawAnything-Sim tasks (3 seeds), we snapshot the canvas at environment step 30 and reset the environment to that snapshot at step 60, erasing 3\,s of drawing progress. Only the environment is reset; the prompt and each model's context are untouched, so ICRT's retained history now contains states that no longer match the current canvas progress. Both models execute actions with a horizon of 10 so that the disturbance lands at the same time step (the default model horizon of 12 used for the main experiments does not divide 30, which is why \ours's undisturbed error differs slightly from Fig.~\ref{fig:main_result}a). Tab.~\ref{tab:disturbance} reports the results; the $\Delta$ error bars are standard deviations of the per-seed paired differences.

\begin{table}[H]
\centering
\begin{tabular}{lccc}
\hline
\textbf{Model} & \textbf{Undisturbed} & \textbf{Disturbed} & \textbf{$\Delta$ error} \\
\hline
\ours & $3.35 \pm 0.09$ & $8.83 \pm 0.25$ & $+5.48 \pm 0.15$ \\
ICRT & $5.07 \pm 0.22$ & $15.54 \pm 0.90$ & $+10.47 \pm 0.98$ \\
\hline
\end{tabular}
\vspace{1mm}
\caption{\textbf{Mid-rollout disturbance on DrawAnything-Sim.} Drawing error on 50 unseen tasks (mean $\pm$ std over 3 seeds) without and with the canvas reverted mid-rollout, and the resulting increase. For disturbed rollouts the error is the lowest reached after the revert.}
\label{tab:disturbance}
\end{table}

\section{Laundry folding details}
\label{sec:appendix_laundry}
This section provides further details for the laundry folding experiment in \S\ref{sec:laundry_folding}.

\textbf{Tasks.} The three tasks are \textit{fold left arm} (requires only the left robot arm), \textit{fold right arm} (requires only the right robot arm), and \textit{fold bottom up} (requires both arms to individually grab the bottom of the sweater and then jointly fold it to the top). All tasks start with the sweater fully unfolded.

\textbf{Data.} Using bimanual iPhUMI, we collect $\sim$150 full demonstrations per task and $\sim$1000 total error correction demonstrations. We collect five new behavior prompts per task for evaluation, so the evaluations in Fig.~\ref{fig:laundry_eval} use \textit{unseen} prompts for \textit{seen} tasks. The \lang baseline is detailed in \S\ref{sec:appendix_baseline_details}.

\textbf{Failure modes.} The \textit{fold bottom up} task first grabs the bottom with the right arm and then with the left arm; we often observe failures where \ours correctly grasps the bottom with the right arm, but the left arm instead decides to fold the left sleeve instead of also grasping the bottom. Given the low task diversity, the indirectness of the prompt as a task descriptor might make \ours sensitive to overfitting to spurious task cues in the training data (e.g., background variations more often seen for a particular task).

\textbf{Outlook.} As expected, behavior prompting performs poorly in this low task diversity regime. In a high task diversity setting, however, we envision behavior prompting having substantial advantages with training data spanning \textit{many} folding styles across \textit{many} garments. In particular, a behavior prompt captures the \textit{temporal} steps in the folding process as well as the \textit{spatial} interaction points, both of which clarify and inform the desired execution in a high task diversity setting. This information could also help with visual adaptation to unseen garments and temporal adaptation to unseen folding orders. Practically, a user might also find it more natural to describe their desired folding preferences via a single demonstration, rather than through language.

\section{BPP Training details}
\label{sec:appendix_training_details}

\textbf{Task groupings.} During training, the prompt and the target observation-action pairs are sampled from the same task grouping. These groupings define the granularity at which behavior prompts can influence robot execution: for example, a pick-place task with two distinct grasping strategies for one object would require two separate task groupings for the robot to adhere to the grasping strategy shown in the prompt.

\begin{table}[H]
\centering
\begin{tabular}{lcc}
\hline
\textbf{Benchmark} & \textbf{Duration (hrs)} & \textbf{GPUs} \\
\hline
\liberogencombination & 28 & 4$\times$ NVIDIA RTX A6000 \\
\liberogenchain       & 34 & 4$\times$ NVIDIA L40S \\
DrawAnything-Sim      & 27 & 4$\times$ NVIDIA L40S \\
DrawAnything-Real     & 42 & 4$\times$ NVIDIA L40S \\
Laundry Folding       & 24 & 4$\times$ NVIDIA L40S \\
\hline
\end{tabular}
\vspace{1mm}
\caption{\textbf{Compute requirements for training BPP for each training benchmark.} Simulation experiments include time for intermediary and final rollout evaluations. We train using DDP.}
\label{tab:compute}
\end{table}

\begin{table}[H]
    \centering
    \begin{tabular}{c c c c c}
        \hline
        Layers & Hidden size $D$ & MLP size & Heads & Params \\
        \hline
        6 & 768 & 3072 & 8 & 57M \\
        \hline
    \end{tabular}
    \vspace{1mm}
    \caption{\textbf{BPP prompt encoder architecture (transformer decoder).}}
    \label{tab:pe_arch_appendix}
\end{table}

\begin{table}[H]
    \centering
    \begin{tabular}{l c c c c}
        \hline
        Structure & Step emb dim & Down dims & Kernel & Params \\
        \hline
        U-Net & 128 & [256, 512, 1024] & 5 & 151M \\
        \hline
    \end{tabular}
    \vspace{1mm}
    \caption{\textbf{Action decoder model architecture details.} We use the CNN U-Net action diffusion architecture from~\citet{chi2023diffusionpolicy}. \ours, \lang, and \goalimage models use this for action diffusion.}
    \label{tab:diffusion_arch_appendix}
\end{table}

\clearpage

\section{iPhUMI}
\label{sec:appendix_iphumi}
\begin{figure}[H]
    \centering
    \includegraphics[width=.9\linewidth]{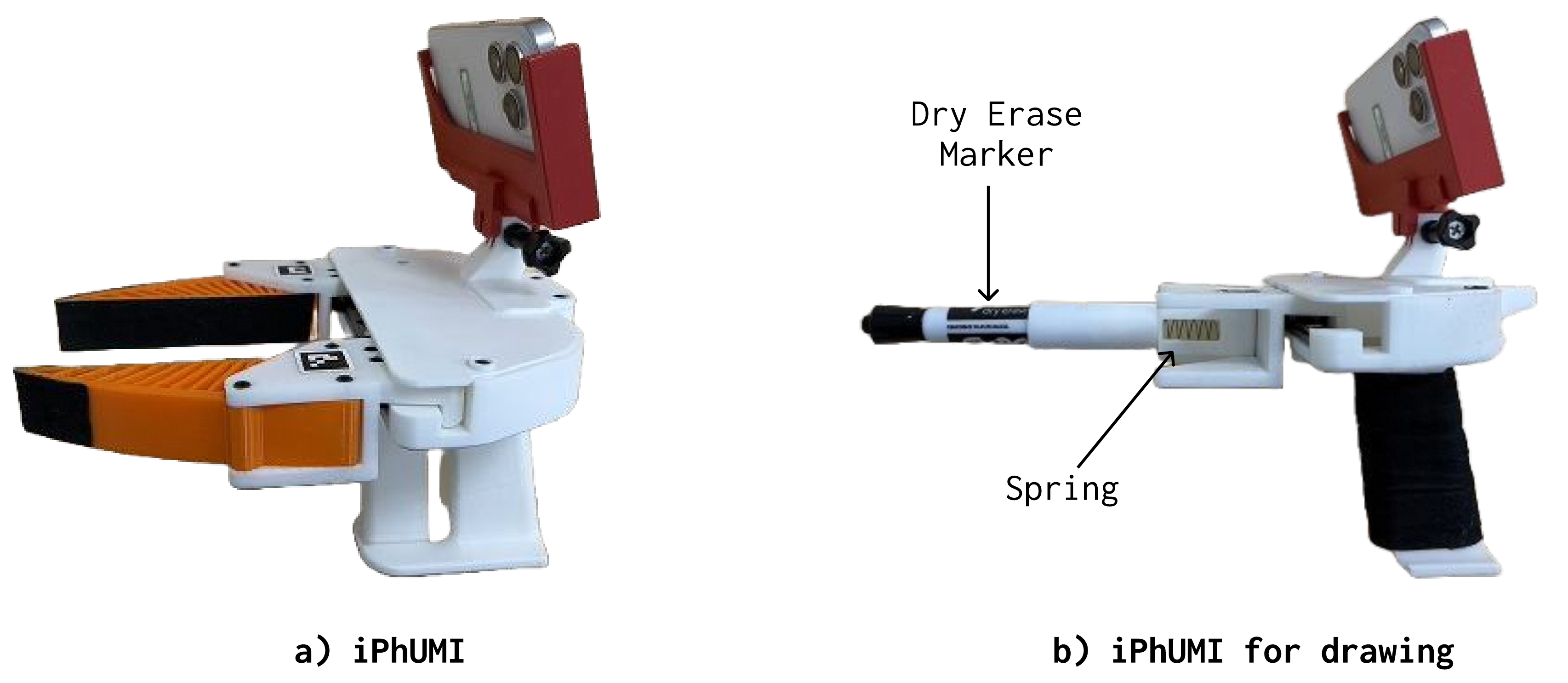}
    \caption{\textbf{The iPhUMI handheld data collection gripper.} iPhUMI enables real-time localization in new environments, which dramatically reduces the setup time required for collecting demonstration data compared to the original UMI~\citep{chi2024universal}. It features a custom-built application that facilitates data collection and policy deployment. With iPhUMI, a user can also specify behavior prompts at test-time to immediately condition behavior prompting policies. We present two different instantiations of iPhUMI: a) with fingers for standard manipulation tasks and b) with a marker attachment and spring for compliant drawing.}
    \label{fig:iphumi}
\end{figure}

\begin{figure}[H]
    \centering
    \includegraphics[width=0.9\linewidth]{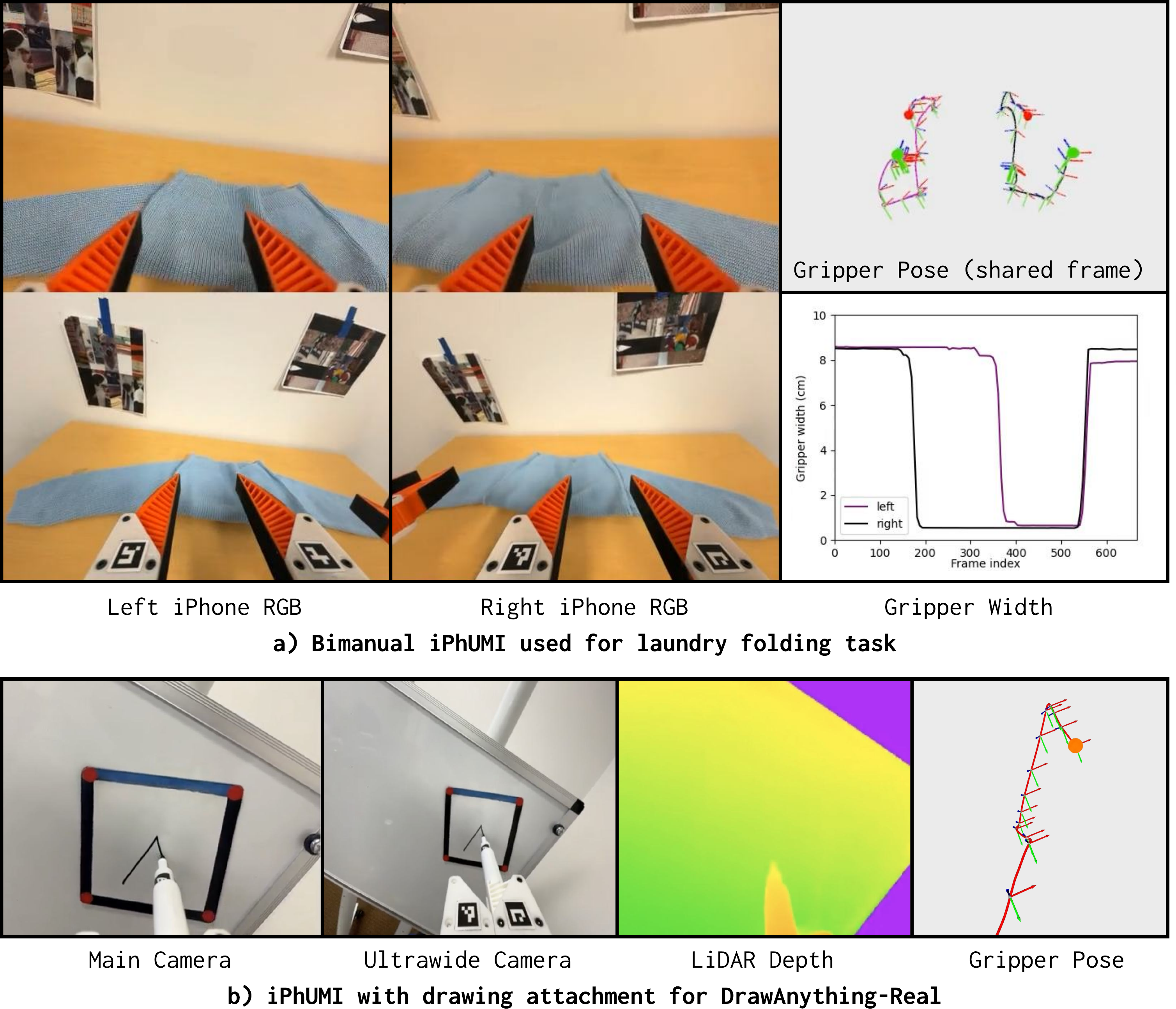}
    \caption{\textbf{iPhUMI collected modalities.} Data collected with bimanual iPhUMI for a) laundry folding and b) drawing visualized part-way through a user drawing the letter A. For DrawAnything-Real we do not use the ultrawide or depth camera as policy inputs, but include them for reference.}
    \label{fig:iphumi-data}
\end{figure}

iPhUMI is a handheld data collection interface designed to enable rapid data collection across new tasks and environments. We modify the original UMI gripper design \citep{chi2024universal} by replacing the GoPro with a custom 3D-printed mount to hold an iPhone (Fig.~\ref{fig:iphumi}a). For drawing tasks, the UMI gripper is further modified to attach a marker with a spring to allow compliance for drawing without the need for a force-torque sensor (Fig.~\ref{fig:iphumi}b). We open source a custom iOS application for iPhUMI that provides an intuitive application interface for recording demonstration data and for using the iPhone as a camera on the robot during policy deployment.

\subsection{iPhUMI data collection}

\begin{figure}[H]
    \centering
    \includegraphics[width=1\linewidth]{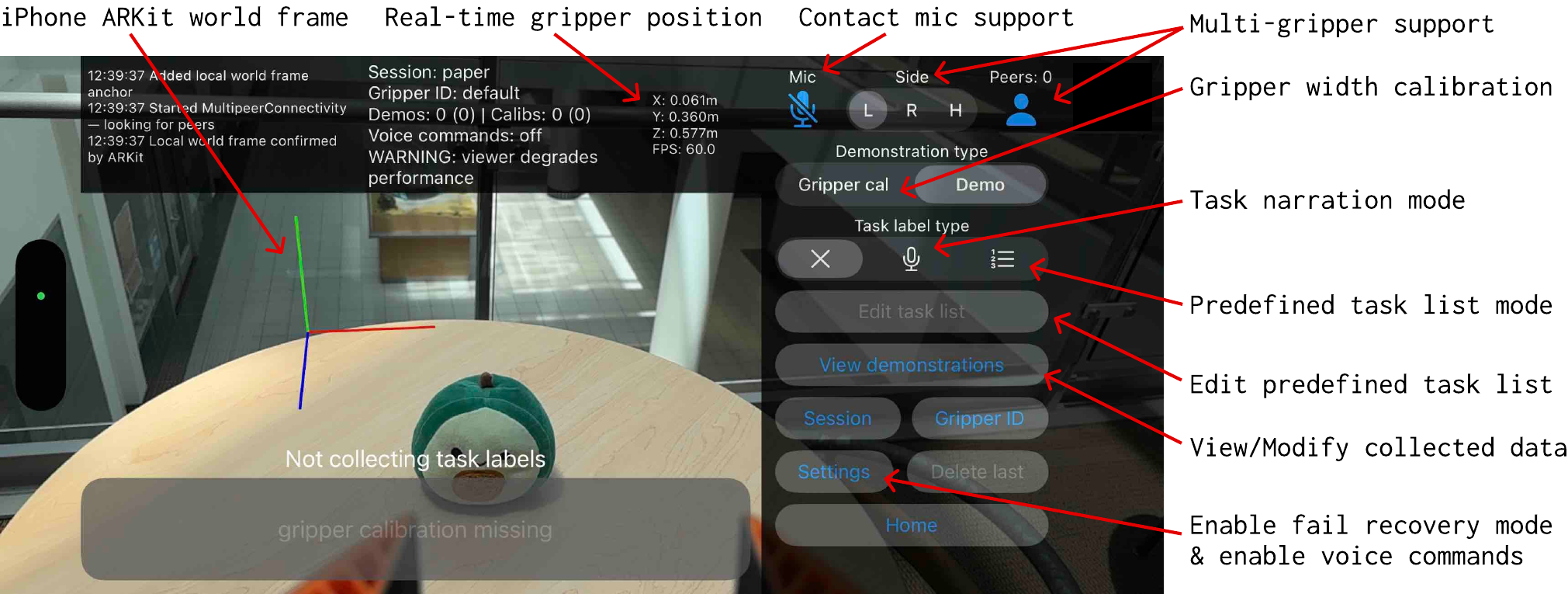}
    \vspace{-4mm}
    \caption{\textbf{iPhUMI data collection interface.} We provide an interface to perform gripper calibration, collect demonstrations, and set appropriate settings for data collection.}
    \label{fig:iphumi-data-collection}
\end{figure}

iPhUMI is capable of collecting five types of data from the iPhone simultaneously: main camera (1920x1440 at 60Hz), ultrawide camera (640x480 at 10 Hz), LiDAR depth (256x192 at 60Hz), gripper pose (60Hz), and gripper width (10Hz) (Fig. \ref{fig:iphumi-data}). Note that depth data is not used in any models or experiments presented in this paper. Unlike previous iPhone-based grippers that either rely on an external fisheye camera~\citep{fang2026robopocketimproverobotpolicies} or do not capture the ultrawide camera~\citep{etukuru2024robot}, our iPhUMI application is capable of capturing the ultrawide camera input. This provides increased visual context to the policy, which has been found to improve performance~\citep{chi2024universal}.

The primary data collection interface is shown in Fig. \ref{fig:iphumi-data-collection}. Upon launching the app, ARKit initializes a world coordinate frame and then continually estimates the iPhone pose using SLAM. The resulting 6-DoF pose of the iPhone in this world frame is logged in real time throughout each demonstration. The app also supports simultaneous recording from multiple grippers. Up to three devices (bimanual iPhUMI and head-mounted iPhone) can be connected into a shared ARKit session to share a single world coordinate frame among them. The triple iPhone setup was co-developed with~\citet{xu2026hommi}. The iPhones will connect into a shared session once common world features have been detected between them.

Before collecting demonstrations, the user performs a one-time gripper width calibration by recording a $\sim$10\,s clip while opening and closing the gripper multiple times in calibration mode. The iPhUMI uses the ultrawide camera to detect ArUco tags on the fingers to determine the minimum and maximum gripper width for the specific hardware.

When collecting data, the user can specify the task name(s) associated with the demonstration and switch labels online as the task changes. Alternatively, the user may narrate the task while demonstrating the task, and iPhUMI uses speech-to-text to label segments with the active task. We also have support for connecting the contact microphone used by~\citep{liu2025maniwav} through the USB-C connector on the iPhone to capture synchronized audio data.

\begin{figure}[H]
    \centering
    \includegraphics[width=0.85\linewidth]{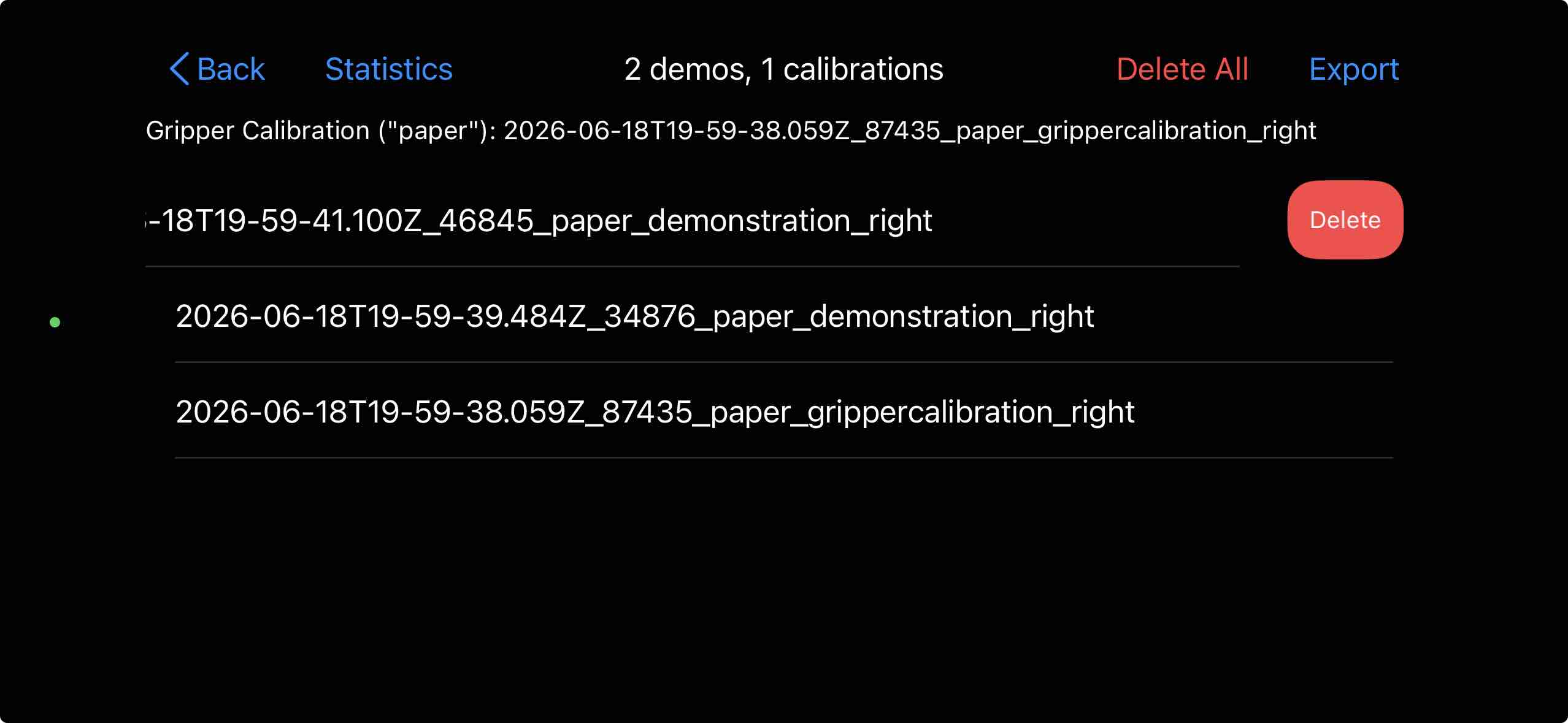}
    \vspace{-1mm}
    \caption{\textbf{iPhUMI Demonstration management interface.} This interface lets the user view and manage collected demonstration data. The data can also be exported to an external SD card connected with a USB C adapter.}
    \label{fig:iphumi-demo-view}
\end{figure}

Recorded demonstrations can be reviewed using the interface shown in Fig \ref{fig:iphumi-demo-view}. Selecting a demonstration displays the main, ultrawide, and depth videos, while a left swipe allows the user to delete a demonstration. Tapping the \texttt{Export} button saves the demonstration data to an external location such as an SD card connected via a USB C adapter.

\subsection{iPhUMI policy deployment}

\begin{figure}[H]
    \centering
    \includegraphics[width=0.85\linewidth]{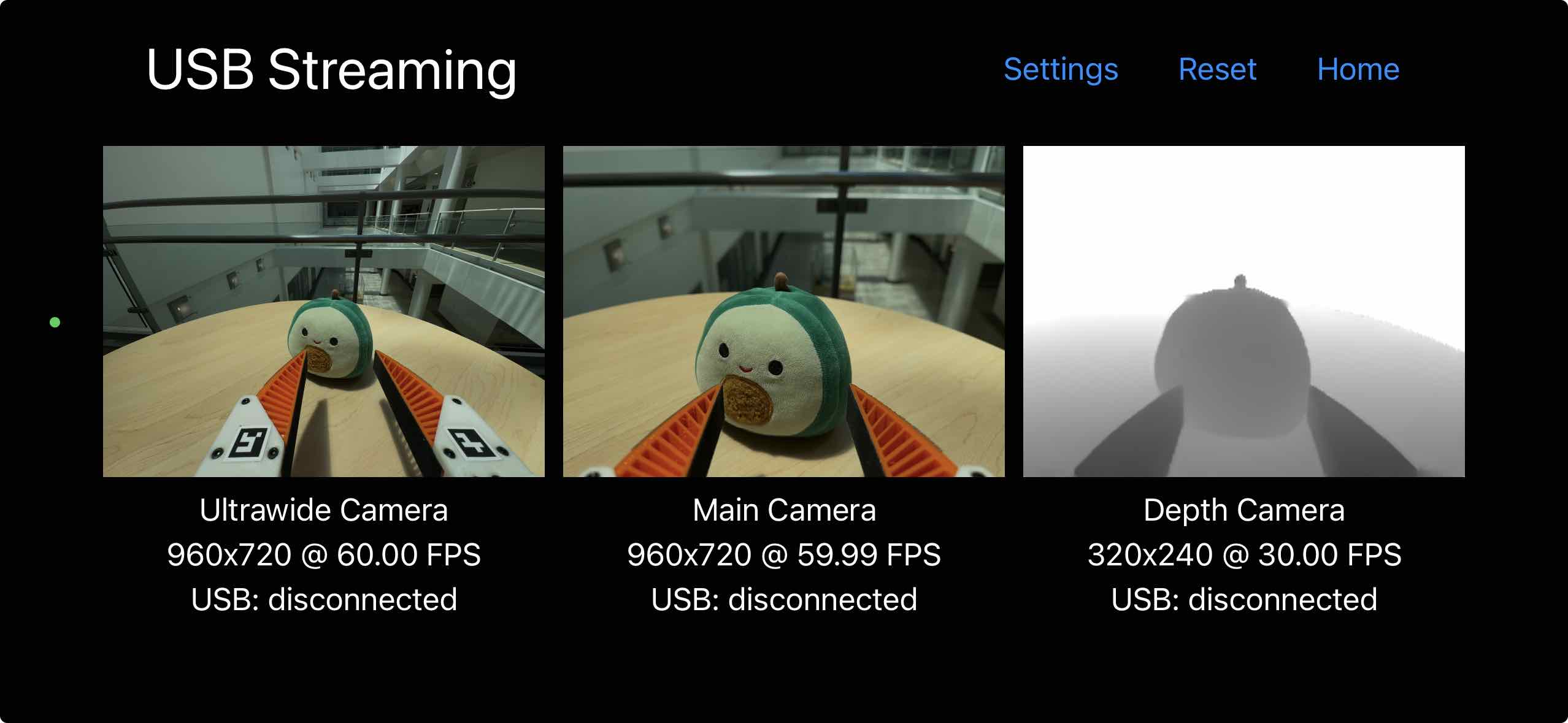}
    \vspace{-1mm}
    \caption{\textbf{iPhUMI deployment interface}. Both USB and Ethernet streaming are supported to stream main camera, ultrawide camera, and LiDAR depth for use during robot deployment.}
    \label{fig:iphumi-deployment}
\end{figure}

The deployment interface (Fig. \ref{fig:iphumi-deployment}) can be used to deploy a policy on the robot using an iPhone as the camera. The iPhone streams its camera feeds and depth through a USB or Ethernet connection. The main camera and ultrawide camera feeds are each streamed 960x720 at 60Hz. Depth is streamed at 320x240 at 30Hz. Our robot deployment infrastructure is built off the code from~\citep{gao2026gatedmemorypolicy}.

\subsection{iPhUMI for test-time behavior prompting}

The iPhUMI app supports wireless transfer of behavior prompts from the iPhone to a deployment desktop connected to a robot. This enables rapid behavior prompt collection and conditioning of robot execution with \ours. When in behavior prompting mode, a user collects a single demonstration using iPhUMI that is then wirelessly transmitted to a desktop connected to a robot. The desktop processes the raw iPhone data into a behavior prompt that is used to condition \ours. This enables a rapid process for practically leveraging behavior prompting at test-time to specify new tasks.

\end{document}